\documentclass[journal,onecolumn]{IEEEtran}
\usepackage{array,multirow} 
\usepackage{cite}
\usepackage{algorithm}
\usepackage{slashbox,pict2e}
\newcommand{\tabincell}[2]{\begin{tabular}{@{}#1@{}}#2\end{tabular}}
\ifCLASSINFOpdf
  \usepackage[pdftex]{graphicx}
\else
  \usepackage[dvips]{graphicx}
\fi
\usepackage{algorithmic}
\begin{document}
\title{Subpopulation Diversity Based Selecting Migration Moment in Distributed Evolutionary Algorithms}
\author{Chengjun~Li and Jia~Wu,~\IEEEmembership{Member,~IEEE}
\thanks{C. Li is with the School of Computer Science, China University of Geosciences, Wuhan 430074, China, and the Centre for Quantum Computation and Intelligent Systems, Faculty of Engineering \& Information Technology, University of Technology Sydney, Australia (e-mail: chengjun.li@uts.edu.au.)}
\thanks{J. Wu is with Centre for Quantum Computation and Intelligent Systems, Faculty of Engineering \& Information Technology, University of Technology Sydney, Australia (e-mail: jia.wu@uts.edu.au)}
}

\markboth{}%
{Shell \MakeLowercase{\textit{et al.}}: Bare Demo of IEEEtran.cls for Journals}
\maketitle
\begin{abstract}
In distributed evolutionary algorithms, migration interval is used to decide migration moments.
Nevertheless, migration moments predetermined by intervals cannot match the dynamic situation of evolution.
In this paper, a scheme of setting the success rate of migration based on subpopulation diversity at each interval is proposed.
With the scheme, migration still occurs at intervals, but the probability of immigrants entering the target subpopulation will be determined by the diversity of this subpopulation according to a proposed formula.
An analysis shows that the time consumption of our scheme is acceptable.
In our experiments, the basement of parallelism is an evolutionary algorithm for the travelling salesman problem.
Under different value combinations of parameters for the formula, outcomes for eight benchmark instances of the distributed evolutionary algorithm with the proposed scheme are compared with those of a traditional one, respectively.
Results show that the distributed evolutionary algorithm based on our scheme has a significant advantage on solutions especially for high difficulty instances.
Moreover, it can be seen that the algorithm with the scheme has the most outstanding performance under three value combinations of above-mentioned parameters for the formula.
\end{abstract}

\IEEEpeerreviewmaketitle
\section{Introduction}
\IEEEPARstart{A}{s} stochastic search methods based on population, evolutionary algorithms (EAs) have been applied successfully in many fields~\cite{alba2002parallelism,wu2013artificial}.
Nevertheless, for nontrivial problems, EAs tend to premature convergence.
Fortunately, using a parallel EA (PEA) often leads to a superior numerical performance because EAs are naturally prone to parallelism~\cite{alba2002parallelism}.
Among types of PEAs, distributed EAs (DEAs), which can be readily implemented in distributed memory MIMD computers~ \cite{alba2002parallelism}, are most popular~\cite{alba2002parallelism,cantu1998survey}.

In a DEA, a large population is divided into a number of subpopulations.
The parallel operator, migration,
exchanges individuals between subpopulations at intervals~\cite{herrera1999hierarchical}.
As a result, global convergence behavior is improved.
The first idea of DEA can be traced back to \cite{bossert1967mathematical}.
So far, many papers focus on DEAs.
In recent years, the application range of DEA is wider and wider for more and more suitable parallel computing environments being available.
For instance, in \cite{tosun2013robust,zidi2013distributed,bulnes2013parallel,rocha2014hybrid,devos2014simultaneous}, there are DEAs for problems of different fields.
On such a background, it is meaningful to improve migration operator for enhancing DEAs.
As one way for this purpose, diversity based migration has been studied in some papers (e.g. \cite{denzinger2003improving,power2005promoting,wei2009maintain,araujo2011diversity}).

The main motivation of this paper is as follows.
According to literature, diversity based migration do help DEAs to get better solutions.
However, existing related studies only focus on selecting fit individuals for migration.
In fact, diversity based migration may also be realized by selecting migration moments, which is determined only by the parameter, migration interval in traditional DEAs.
Nevertheless, such a topic is rarely seen in literature.

A scheme of setting the success rate of migration based on subpopulation diversity at each interval is proposed in this paper for DEAs.
The idea behind it is selecting migration moments based on not only interval but also subpopulation diversity.
More precisely, migration still occurs at intervals, but the probability of immigrants entering the target subpopulation, $p$, will be decided by the diversity of this subpopulation, $d$, according to a formula building the relationship of them.
In detail, at migration intervals, $d$ is computed first.
Then, $p$ is calculated according to the formula.
At the probability, $p$, immigrants enter the subpopulation .

An experiment is conducted on eight instances of the Travelling Salesman Problem (TSP) from \cite{reinelt1991tsplib} to compare the outcomes of the DEA with the proposed scheme and those of a traditional DEA.
The only difference between them is that the proposed scheme is used only in the former.
For the formula in the scheme, two parameters, $\alpha$ and $\beta$, are required.
In our experiment, the common parameters in both algorithms are set the same value.
Moreover, the algorithm based on the scheme is tested under nine value combinations of $\alpha$ and $\beta$.
Results show that the DEA based on our scheme can get better and more stable solutions than the traditional one.
Further, under some value combinations of $\alpha$ and $\beta$, the advantage of the algorithm with our scheme is much more remarkable.

The rest of this paper is organized as follows.
Related researches are described in Section 2.
In Section 3, the proposed scheme is presented.
Then, experimental results are shown and analyzed in Section 4.
Finally, a conclusion and a prospect are dealt with in Section 5.
%
\section{Related Researches}
\subsection{Distributed Evolutionary Algorithms}
A DEA can be considered as the upgrade of a EA to enhance solving ability.
In \cite{herrera1999hierarchical}, it is expressed as the Algorithm~\ref{alg1} whose \emph{italic part} is the migration operation.
\begin{algorithm}
\caption{Traditional DEA}
\label{alg1}
\begin{algorithmic}[1]
\STATE Generate at random a population, $P$, of individuals
\STATE Divide $P$ into $SP_1$, \ldots, $SP_n$ subpopulations
\STATE Define a neighborhood structure for $SP_x$, $x=1$, \ldots, $n$
\REPEAT [Execute in parallel the next steps for $SP_x$, $x=1$, \ldots, $n$]
  \STATE Apply, during $i$ generations, stochastic operators of EA
  \STATE \emph{Send $s$ individuals to neighboring subpopulations}
  \STATE \emph{Receive $s$ individuals from neighboring subpopulations}
\UNTIL{the stop criterion is not fulfilled}
\end{algorithmic}
\end{algorithm}
Main parallel parameters in DEAs are introduced as below:
\begin{itemize}
  \item
  Migration topology (\lq\lq neighborhood structure\rq\rq  in Algorithm~\ref{alg1}), it describes which subpopulations send individuals to which subpopulations~\cite{skolicki2005influence}.
  \item
  Migration strategy, it defines how to select emigrants and choice replaced individuals in each subpopulation.
  \item
  Migration rate (\lq\lq r\rq\rq  in Algorithm~\ref{alg1}), it specifies the quantity of individuals which emigrate from each subpopulation in one migration round.
  \item
  Migration interval (\lq\lq i\rq\rq  in Algorithm~\ref{alg1}), it is the every certain number of generations between a migration round and the next one~\cite{skolicki2005influence}.
  \item
  Subpopulation size, it is the quantity of individuals in each subpopulation.
  \item
  Quantity of subpopulation (\lq\lq n\rq\rq  in Algorithm~\ref{alg1}), it is the number of subpopulations.
\end{itemize}
\par
On one hand, semi-isolation can maintain difference among subpopulations.
On the other hand, for subpopulations, at intervals, migration is expected to afford individuals whose fitness are close to that of local ones but have some different building blocks~\cite{cohoon1987punctuated}.
After migration, immigrants participating in variation operations can help to resist premature convergence if they do have some new building blocks.
In this way, stagnation is postponed.
\subsection{Diversity Based Migration in DEAs}
Diversity refers to differences among individuals, which can be measured at the genotype or phenotype levels \cite{vcrepinvsek2013exploration}.
Thus, diversity of a colony shows the its convergence degree.
Though diversity measures at the latter level often demands much less time consumption, they are not as accurate as ones at the former one because the difference on fitness is not equivalent to that on chromosome structure.

Some instances on diversity based migration from literature are listed as below.
Their diversity measures are all at the genotype level.
In \cite{denzinger2003improving}, a weighted sum of fitness and diversity measure is used in a DEA for selecting emigrants and replaced individuals.
Results showed a reduction in the time taken to find the optimal solution across a range of benchmark problems.
In the DEA presented by \cite{power2005promoting}, the subpopulation representative what is an individual having the lowest average of distance to other all individuals in this subpopulation is found by calculation.
Then, it is selected to be the emigrant.
Furthermore, if migration rate is set more than one, some other individuals are chosen as emigrants based on the distance from it.
This strategy has been tested on a number of problems and has consistently outperformed the standard migration strategy.
The scheme in \cite{wei2009maintain} only allows different enough individuals migrate to a subpopulation in order to keeping diversity of this subpopulation.
To realize this scheme, a method for evaluating the similarity of two subpopulations is presented.
More recently, multikulti methods, are proposed in \cite{araujo2011diversity} for selecting individuals in source subpopulation different enough to ones in target subpopulation .
In order to do this, information on the composition of source subpopulation is required when emigrants are selected.
Different ways of providing this information in a concise manner are considered.
The results of experiments prove the usefulness of the multikulti strategies.
Besides, The success of this kind of strategies is explained via the measurement of entropy as a representation of population diversity.
In conclusion, existing schemes for diversity based migration concentrate on selecting individuals for migration with the guide of diversity.

\section{Proposed Scheme Description}
\subsection{Preliminary}
Migration may bring new building blocks for target subpopulation, while increases the similarity between source subpopulation and target one inevitably.
Hence, there are many occasions not fit for migration during a run.
For instance, in a subpopulation, migration may not be required when getting better individuals by variation operators is easy.
In such a case, migration have little positive influence but still decreases the difference between subpopulations.
Anyhow, migration moments predetermined by intervals cannot match the dynamic situation of evolution.
Consequently, inappropriate migration cannot be avoided effectively.
\subsection{Scheme to Solve Existing Problem}
%
Our thinking is that more migration chances should be provided when subpopulation diversity is low.
In general, the computation for subpopulation diversity is time-consuming in most EAs.
Consequently, it is not fit to decide migration moments only based on subpopulation diversity since, in such a plan, the computation for subpopulation diversity has to be executed frequently.
Also, setting a value of subpopulation diversity as the threshold for migration is not adopted by us.
Two of reasons are listed below.
Firstly, the fitting value is difficult to find and may vary much from one occasion to another.
Besides, migration may not occur at all in some periods, such as the initial stage.

In our scheme, the attempt of migration still occurs at intervals, but the probability of immigrants entering the target subpopulation, $p$, will be computed according to the formula that builds the relationship between subpopulation diversity, $d$, and the probability, $p$.
At first, the simplest formula to describe the relationship between $p$ and $d$, Formula~\ref{eq:01}, is designed by us.
\begin{equation}
  p =1 - d
\label{eq:01}
\end{equation}
In the initial stage of a run, the value of $d$ is close to the maximum, 1.
According to Formula~\ref{eq:01}, $p$ is near to 0 at this time.
Then, $d$ decreasing monotonically makes $p$ increasing monotonically.
In the end, $p$ approaches 1 since $d$ has come to the minimum, 0.
Then, Formula~\ref{eq:02}, where $\alpha\in[0,+\infty]$, and Formula~\ref{eq:03}, $\beta\in[0,+\infty]$ are considered.
\begin{equation}
  p =1 - d^\alpha
\label{eq:02}
\end{equation}
\begin{equation}
  p =(1 - d)^\beta
\label{eq:03}
\end{equation}
Finally, Formula \ref{eq:04} is adopted since the three former ones can be regarded as special cases of it.
\begin{equation}
  p =(1 - d^\alpha)^\beta
\label{eq:04}
\end{equation}
Essentially, $\alpha$ and $\beta$ are just used to adjust the relationship between $p$ and $d$.

The flowchart of a DEA with our scheme is shown as Algorithm~\ref{alg2}.
The scheme is in \underline{\emph{the underlined and italic}} text of Algorithm~\ref{alg2}.
\begin{algorithm}
\caption{DEA with proposed scheme}
\label{alg2}
\begin{algorithmic}[1]
\STATE Generate at random a population, $P$, of individuals
\STATE Divide $P$ into $SP_1$, \ldots, $SP_n$ subpopulations
\STATE Define a neighborhood structure for $SP_x$, $x=1$, \ldots, $n$
\REPEAT [Execute in parallel the next steps for $SP_x$, $x=1$, \ldots, $n$]
 \STATE Apply, during $i$ generations, stochastic operators of EA
 \STATE \emph{Send $s$ individuals to neighboring subpopulations}
 \STATE \emph{Receive $s$ individuals from neighboring subpopulations}
 \STATE \underline{\emph{Compute $d$}}
 \STATE \underline{\emph{Compute $p$ according to the formula}}
 \STATE \underline{\emph{Generate a random number, $r \in [0,1]$}}
 \IF{\underline{$r<p$}}
  \STATE \underline{\emph{Put received individuals into the subpopulation}}
 \ELSE
  \STATE \underline{\emph{discard them}}
 \ENDIF
\UNTIL{the stop criterion is not fulfilled}
\end{algorithmic}
\end{algorithm}
Essentially, our scheme devotes to control the probability of completing migration according to subpopulation diversity at intervals rather than to improve subpopulation diversity.
\par
\subsection{Analysis of Our Scheme on Running Time}
Provided that all supopulations in DEAs remains in exact synchronization, the consuming time for the two algorithms can be expressed as below, respectively.
Let $\Delta t_e$ be the consuming time of all evolutionary operations in one generation, $\Delta t_d$ be that of once diversity computation, $\Delta t_m$ be that of a migration round, $i$ be interval and $g$ be total generations. In a DEA based on the proposed scheme, the total consuming time in a run, $t_w$, can be computed as follow:
\begin{equation}
  t_w = (\Delta t_e + \frac{\Delta t_d + \Delta t_m} {i} ) \times g
\end{equation}
In a DEA without the proposed scheme, $\Delta t'_m$ be the consuming time of a migration round.
Then, the total consuming time in a run, $t_n$ can be computed as follow:
\begin{equation}
 t_n = (\Delta t_e + \frac{\Delta t'_m} {i} ) \times g
\end{equation}
If above two types of DEA are based on the same EA and same in settings of common parameters, comparisons can be done as follow: Firstly,
\begin{equation}
 \Delta t_m \le \Delta t'_m
\end{equation}
mainly because, in the DEA with the scheme, a migration round requires the maximum time, $\Delta t'_m$ , only when $p$ is satisfied.
Moreover, the total consuming time of the former algorithm have a peculiar part,
\begin{equation}
 \frac {\Delta t_d} {i} \times g
\end{equation}
Therefore, the difference of these two algorithms on total consuming time in a run can be measured as below:
\begin{equation}
 \frac {\Delta t_d + \Delta t_m - \Delta t'_m} {i} \times g
\end{equation}
The total consuming time of the former is larger than that of the latter because
\begin{equation}
 \Delta t_d \gg \Delta t'_m - \Delta t_m
\end{equation}
That is, consuming time of once diversity computation is much more than the difference in that of a migration round for different algorithms.

In the scheme, subpopulation diversity is computed only at intervals.
Although the diversity measure may vary according to chromosome coding, the extra consuming time in a DEA for the proposed scheme will always be acceptable if migration interval, $i$, is large enough.
In fact, to maintain the difference between subpopulations, $i$ is always large in the majority of DEAs.
Therefore, this scheme can be widely used in DEAs.
\section{Experiment Studies}
\subsection{Selected Problem and EA for It}
The famous NP-hard problem, TSP, is selected for our experiment.
\cite{reinelt1991tsplib} has its benchmark instances.
Some EA for the TSP have remarkable performance.
For instance, the one presented in \cite{nagata2013powerful} outperforms state-of-the-art heuristic algorithms in finding very high-quality solutions on instances with up to 200,000 cities.

It should be stressed that the purpose of our experiments is not to find better solutions of any problem than ever but to test our scheme by comparing the performance of a DEA with it and that of a traditional one.
For this purpose, the more powerful a EA is, the larger instances, which demand much more on resource, should be used.
In consideration of this, the EA in \cite{cai2005improved} proposed for years is used in our experiment.
It is based on the inver-over operator~\cite{tao1998inver}, the selection method that each individual competes with its offspring only and so-called mapping operator~\cite{cai2005improved}.
The flow of mapping is as follow.
Firstly, two individuals are selected at random.
Then, a segment of chromosome in the individual having worse fitness is selected at random.
After that, a segment which has the same number of cities and the same first city is searched in the other individual.
As soon as it is found, the former segment in the worse individual is replaced by it.
Finally, in the worse individual, the other part of the chromosome is adjusted according to the latter steps of partially mapped crossover~\cite{goldberg1985alleles} which is a traditional operator used in EAs for the TSP.
It can be seen from the flow that mapping belongs to crossover.
In this EA, evolutionary velocity, $v$, which is calculated as the Formula \ref{eq:05}, decides whether the mapping operator should be executed.
\begin{equation}
  {v} = \frac{\left|{{f_b} - {f_b}'} \right|}{\Delta g}
\label{eq:05}
\end{equation}
In this formula, $f_b$ is the fitness of the current best individual, $f'_{b}$ is the that of the previous one and $\Delta g$ denotes the generations between the appearance of the previous best individual and that of the current one.
Mapping is carried out only when $v$ is lower than a threshold value.

\subsection{Diversity Measure in Our DEA}

Matrix M in Formula~\ref{eq:mat} is connection matrix of TSP tour~\cite{chang2010dynamic}.
\begin{equation}
\mathbf{M} =
 \left[{\begin{array}{cccc}
 a_{00} \quad & \quad a_{01}  & \quad \cdots  &  \quad a_{0(k-1)}  \\
 a_{10} \quad & \quad a_{11}  & \quad \cdots  &  \quad a_{1(k-1)}  \\
 \cdots \quad & \quad \cdots  & \quad \cdots  &  \quad \cdots    \\
 a_{(k-1)0} \quad & \quad \cdots &  \quad \cdots & \quad a_{(k-1)(k-1)}
\end{array}} \right]
\label{eq:mat}
\end{equation}
In the matrix, $k$ is the number of cities and $a_{lm} \in \{0,1\}, (0\le l\le k-1,  0\le m\le k-1)$. $a_{lm} = 1$ represents that there is a connection from city\,$(l+1)$ to city\,$(m+1)$ in tour, while $a_{lm} = 0$ denotes that such a connection does not exist.
Let $I_x$ and $I_y$ be two individuals.
Then, each of them has a $k\times k$ connection matrix.
Let $k'$ be the number of rows which are same in the two matrixes.
Then, the difference between $I_x$ and $I_y$ can be defined as Formula~\ref{eq:diff}~\cite{chang2010dynamic}.
\begin{equation}
 D(I_x,I_y) = 1 - \frac{k'}{k}
\label{eq:diff}
\end{equation}
Then, in \cite{chang2010dynamic}, subpopulation diversity is defined as Formula~\ref{form:five}, where $ni$ is subpopulation size and $C$ means combination.
\begin{equation}
 d = \frac{\sum{D(I_x,I_y)}}{C_{ni}^2}
 \label{form:five}
\end{equation}
In this paper, Formula~\ref{form:five} is replaced by Formula~\ref{form:six}  in order to decrease computation complexity.
\begin{equation}
 d = \frac{\sum{D(I_b,I_n)}}{ni-1}
\label{form:six}
\end{equation}
In Formula~\ref{form:six}, $I_b$ is the best individual in subpopulation.
$I_n$ represents each individual in it other than $I_b$.
For once computation of subpopulation diversity, this change makes that total times of the calculation for the difference between two individuals greatly reduce from $C_{ni}^2$ to $ni-1$ in each subpopulation.
Although the average of distance between pairwise individuals is replaced by that from the best individual to another one may lead to some error in the resulting value of subpopulation diversity, convergence degree can still be reflected with much less computation.
As the steps shown in Algorithm~\ref{alg2}, $d$ of each subpopulation is computed at intervals to obtain each $p$.
%
\subsection{Experiment to Compare Result}
\begin{figure}[!htbp]
  \centering
  \includegraphics[width=3.0in]{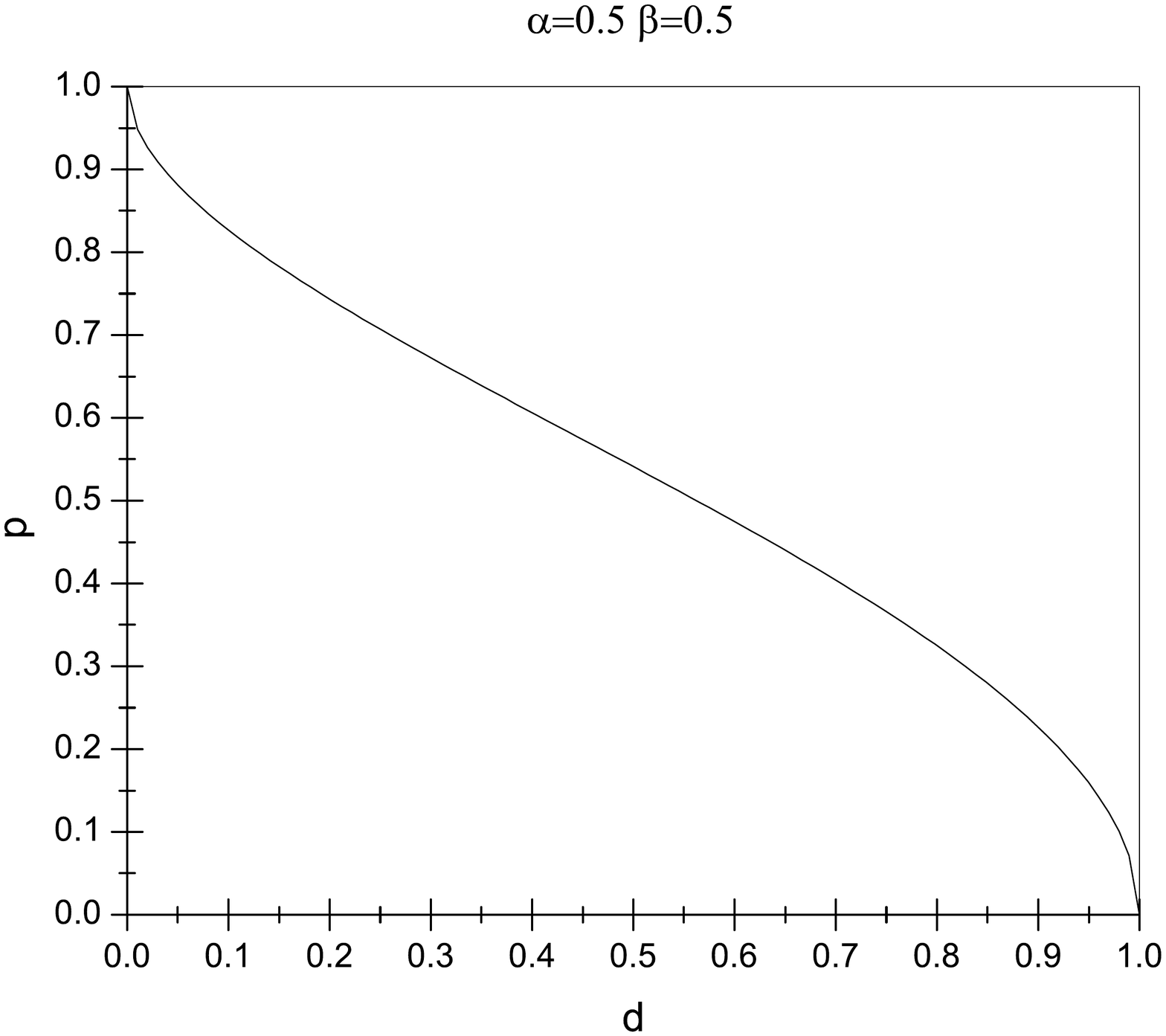}
  \caption{Graph of function for our scheme when $\alpha=0.5$ and $\beta=0.5$ }
  \label{figure:01}
\end{figure}

\begin{figure}[!htbp]
  \centering
  \includegraphics[width=3.0in]{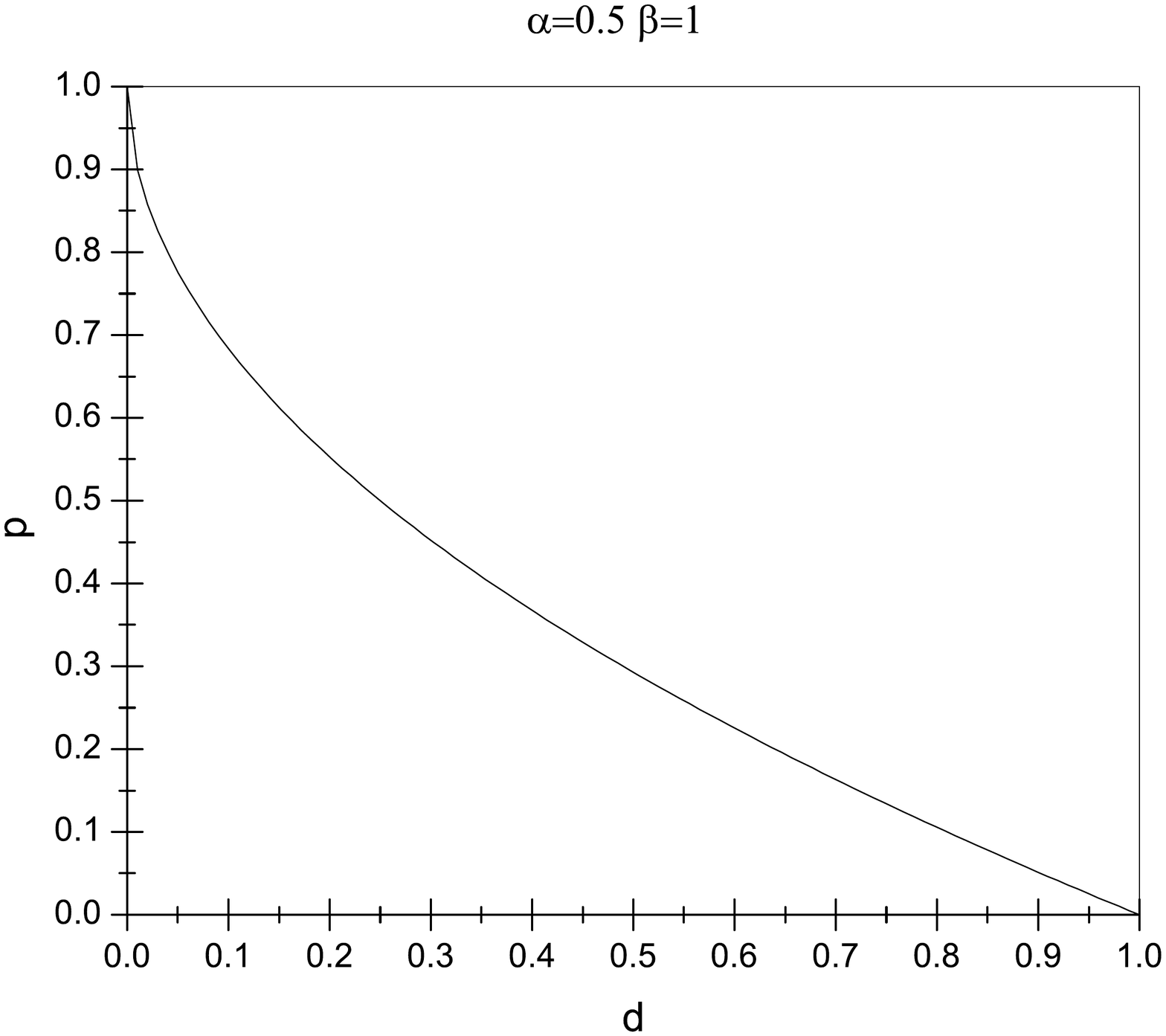}
  \caption{Graph of function for our scheme when $\alpha=0.5$ and $\beta=1.0$ }
  \label{figure:02}
\end{figure}

\begin{figure}[!htbp]
  \centering
  \includegraphics[width=3.0in]{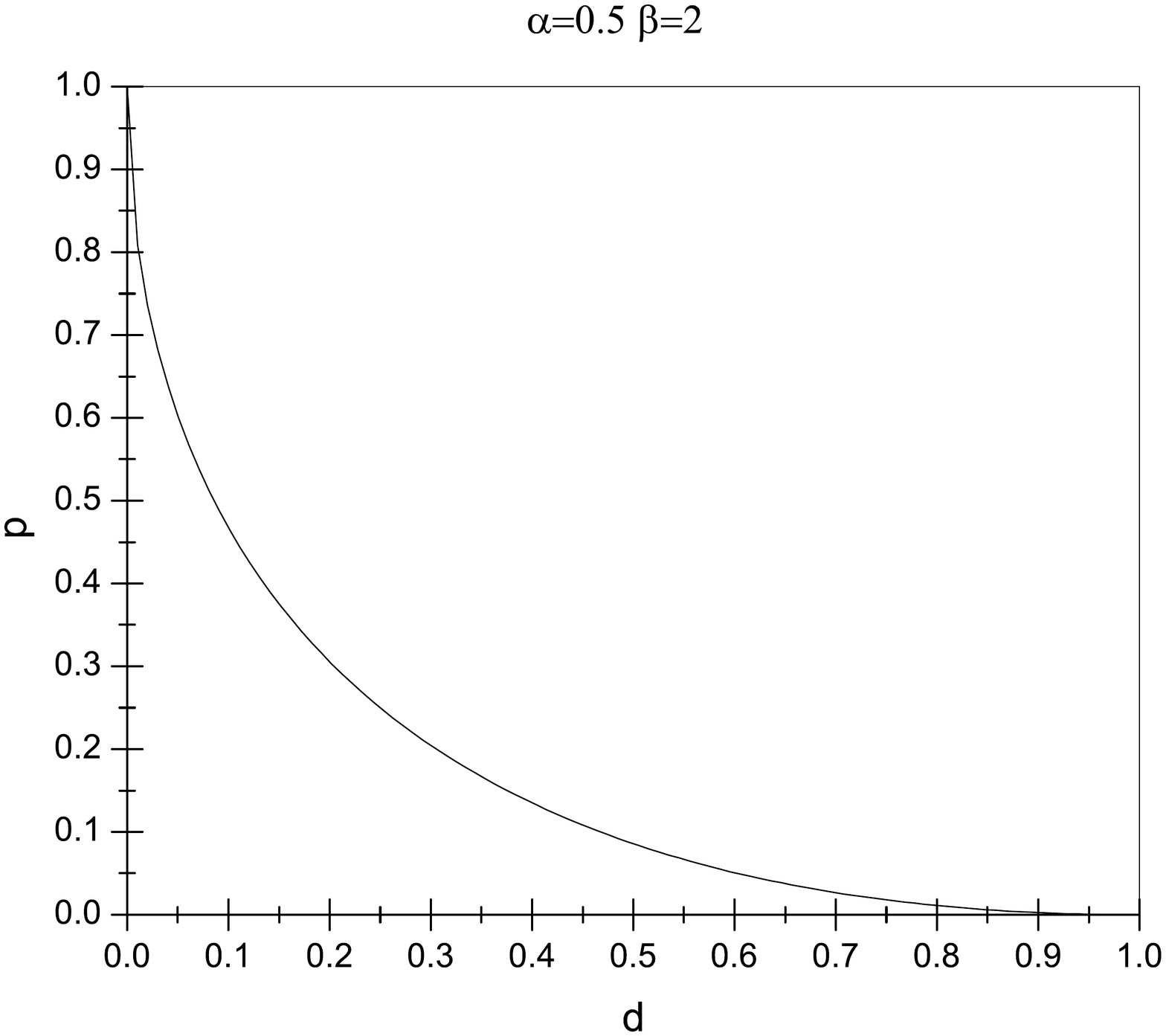}
  \caption{Graph of function for our scheme when $\alpha=0.5$ and $\beta=2.0$ }
  \label{figure:03}
\end{figure}

\begin{figure}[!htbp]
  \centering
  \includegraphics[width=3.0in]{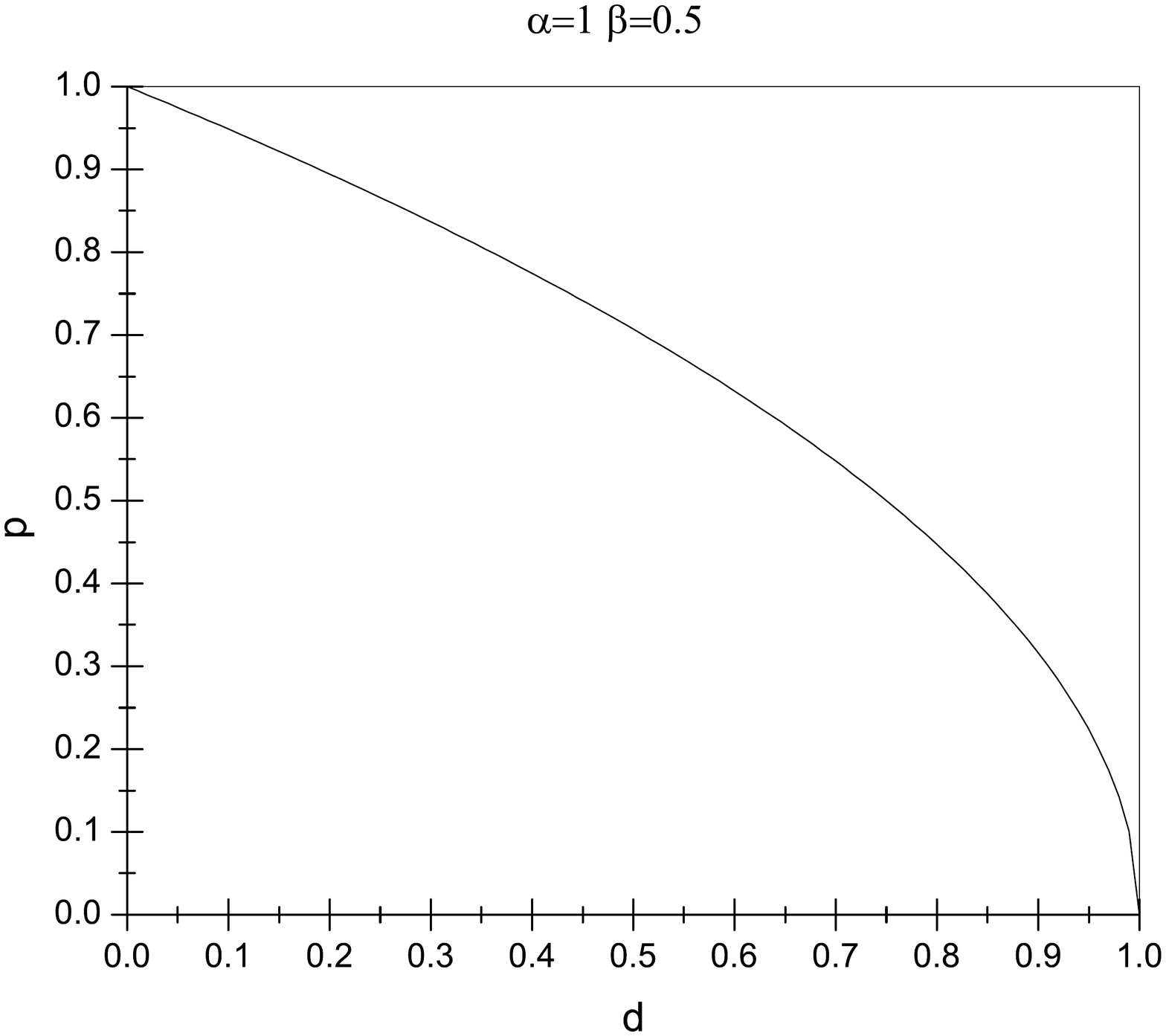}
  \caption{Graph of function for our scheme when $\alpha=1.0$ and $\beta=0.5$ }
  \label{figure:04}
\end{figure}

\begin{figure}[!htbp]
  \centering
  \includegraphics[width=3.0in]{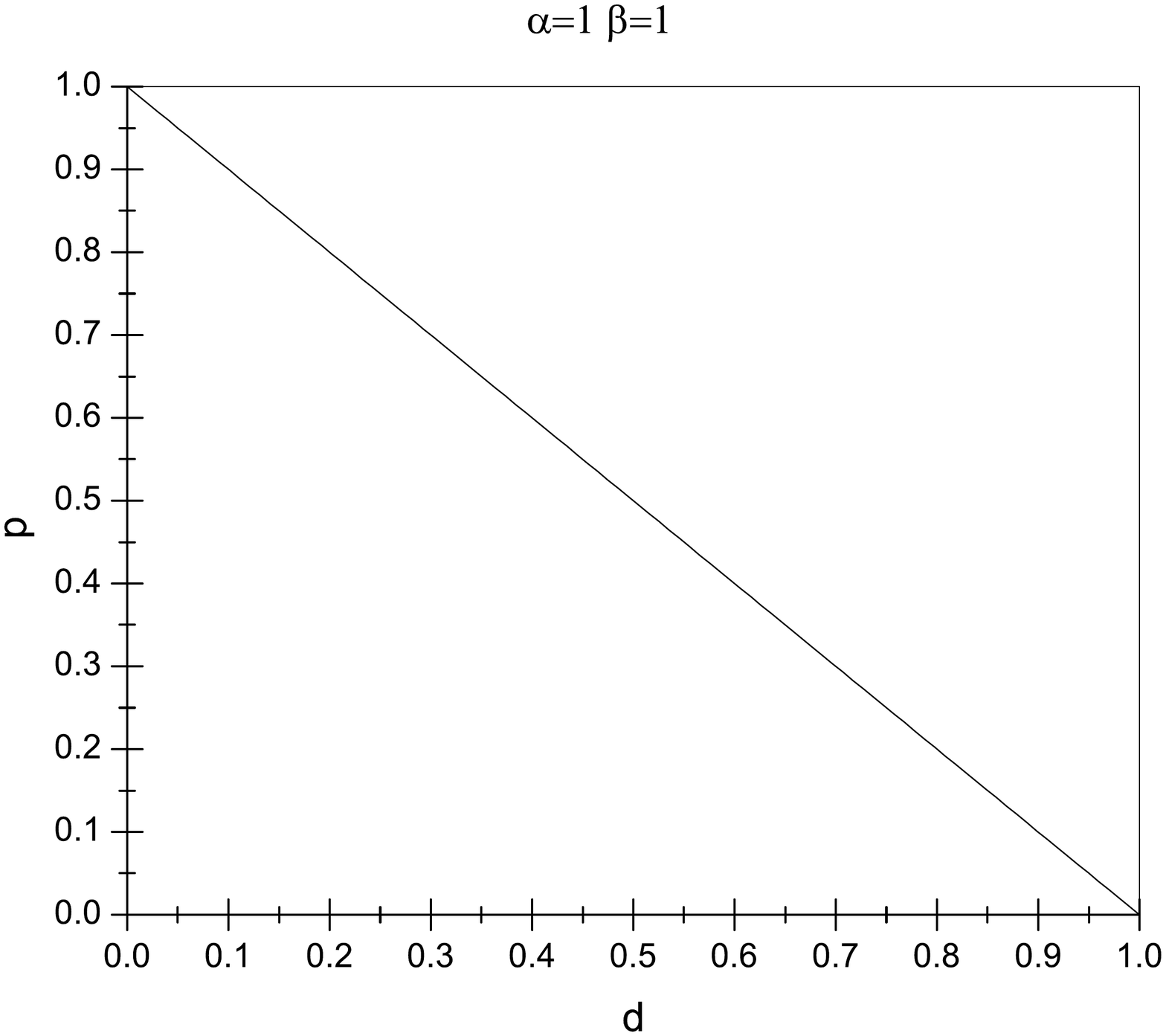}
  \caption{Graph of function for our scheme when $\alpha=1.0$ and $\beta=1.0$ }
  \label{figure:05}
\end{figure}

\begin{figure}[!htbp]
  \centering
  \includegraphics[width=3.0in]{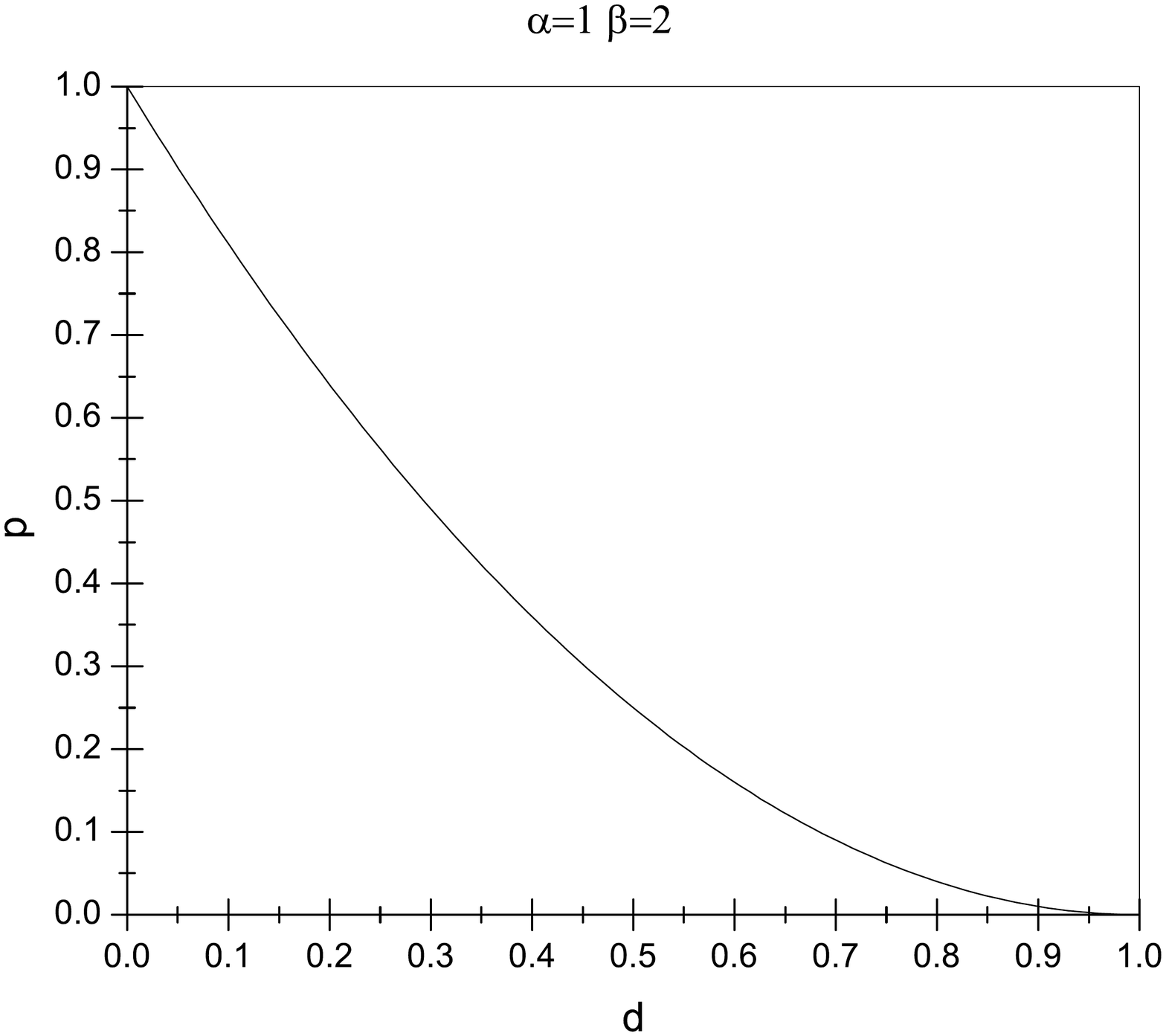}
  \caption{Graph of function for our scheme when $\alpha=1.0$ and $\beta=2.0$ }
  \label{figure:06}
\end{figure}

\begin{figure}[!htbp]
  \centering
  \includegraphics[width=3.0in]{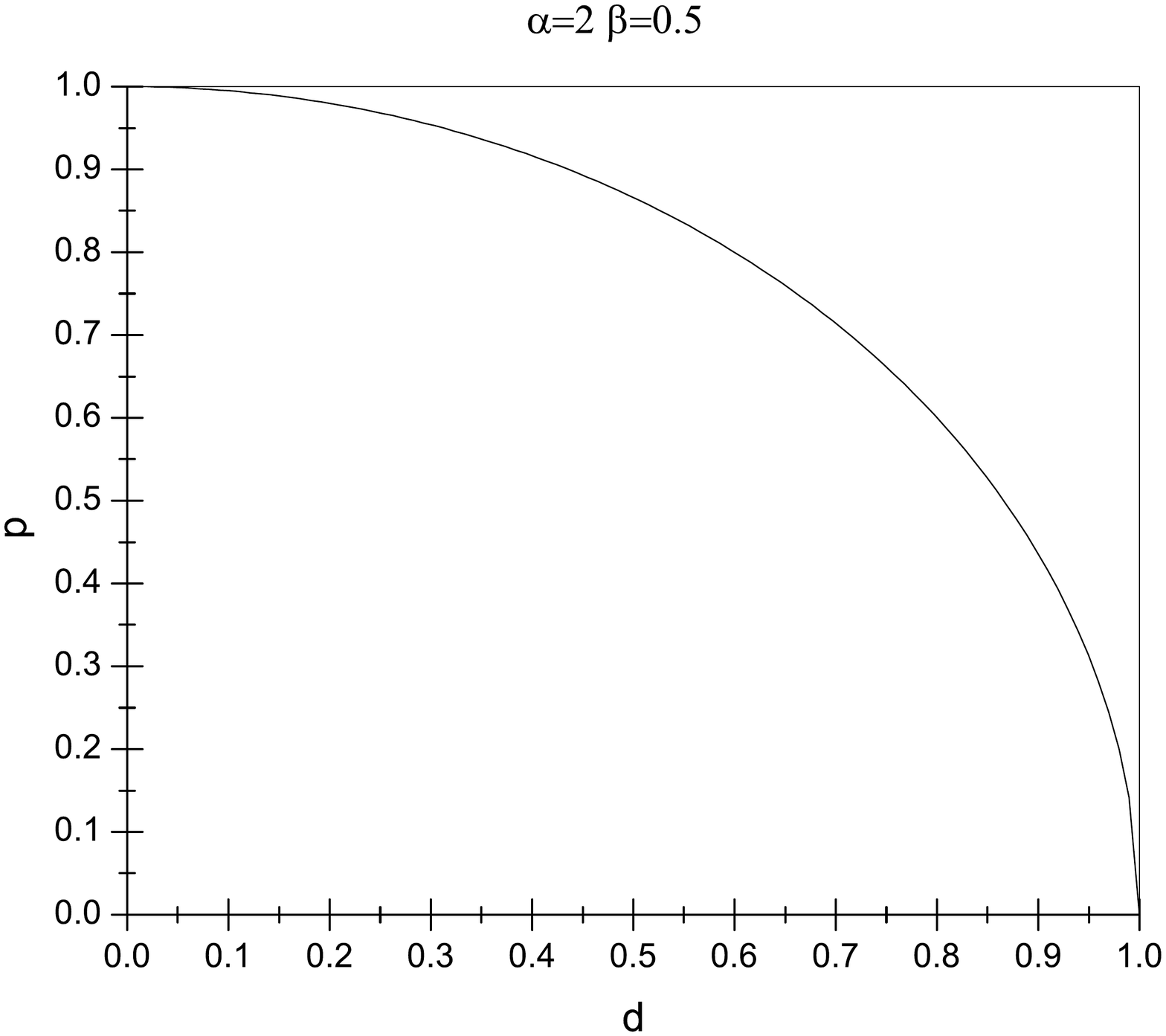}
  \caption{Graph of function for our scheme when $\alpha=2.0$ and $\beta=0.5$ }
  \label{figure:07}
\end{figure}

\begin{figure}[!htbp]
  \centering
  \includegraphics[width=3.0in]{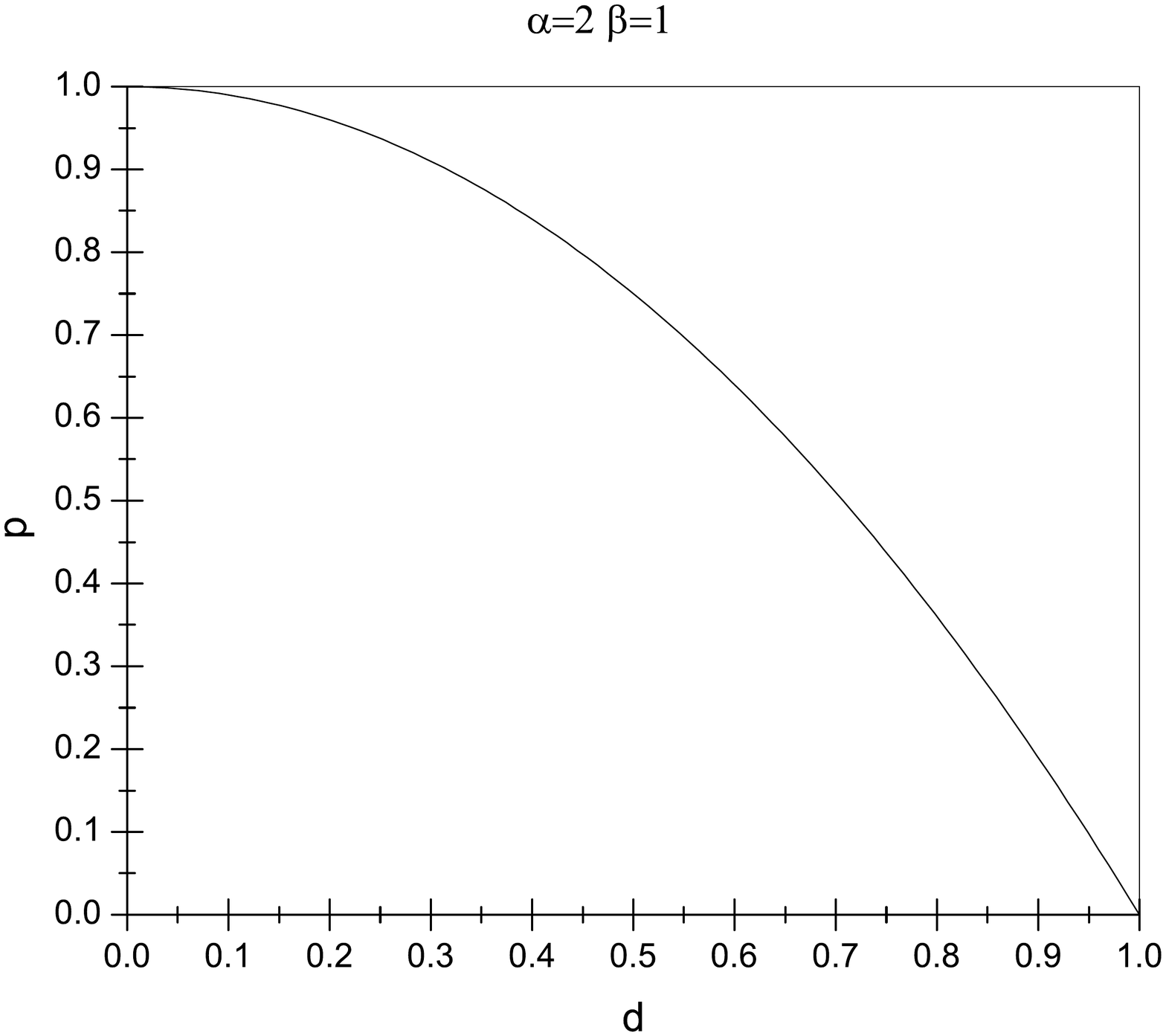}
  \caption{Graph of function for our scheme when $\alpha=2.0$ and $\beta=1.0$}
  \label{figure:08}
\end{figure}

\begin{figure}[!htbp]
  \centering
  \includegraphics[width=3.0in]{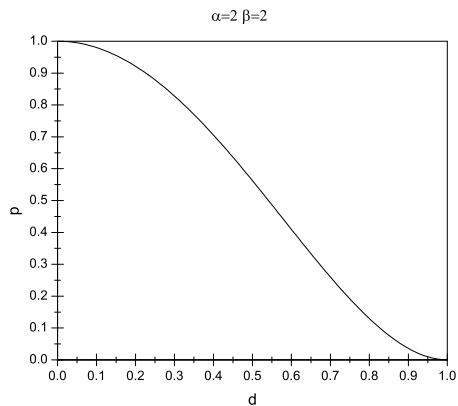}
  \caption{Graph of function for our scheme when $\alpha=2.0$ and $\beta=2.0$ }
  \label{figure:09}
\end{figure}

Our experiment is carried on a Drawing TC5000A computing platform.
It has 1264 2.6\,GHz cores.
Its memory capacity is 1.5\,TB.
In our experiment, $\alpha$ and $\beta$ are set different value combinations to find the fit relationship between $p$ and $d$.
In detail, their values are both get from the set, $\{0.5,1.0,2.0\}$.
In total, there are nine value combinations.
Fig.~\ref{figure:01}-\ref{figure:09} are graphs of the function expressed by Formula~\ref{eq:04} under different value combination of $\alpha$ and $\beta$, respectively.
Then, the outcomes under each value combination are compared with those of a traditional DEA, respectively.
The algorithms in the experiment are both based on the EA introduced in Subsection 4.1 and are the same in setting of common parameters.
The thinking of setting for these common parameters comes from \cite{li2014global}.
In detail, common parameters except migration interval are set the same value in both algorithms.
Then, for each instance, the traditional algorithm runs thirty times independently under five equal difference intervals, respectively.
Under each of the nine value combinations of $\alpha$ and $\beta$, so does the DEA with our scheme.
The value of common parameters except migration interval is listed in Table~\ref{table:01}.
\begin{table}[!htbp]
  \caption{Setting of parameters}
  \centering
  \label{table:01}
  \begin{tabular}{|m{1.4cm}<{\centering}|m{2.5cm} <{\centering}|m{3.6cm}|}
  \hline
  \multirow{4}[20]{2.0cm}{Evolutionary parameters}
          & Mutation rate ($p_{mu}$) & Changing during a run according to Formula~\ref{form:seven} based on the initial value, 0.02.  \\ \cline{2-3}
          & Crossover rate      & 1-$p_{mu}$  \\ \cline{2-3}
          & Mapping rate ($p_{ma}$)   & Changing during a run according to Formula~\ref{form:eight} based on the initial value, 0.05.  \\ \cline{2-3}
          & Threshold value of evolutionary velocity & 5000 \\ \cline{1-3}
           \hline
  \multirow{6}[7]{2.0cm}{Parallel parameters}
          & Migration topology  & Ring  \\ \cline{2-3}
          & Migration strategy  & Random-random  \\ \cline{2-3}
          & Migration size      & 1      \\ \cline{2-3}
          & Subpopulation size  & 100     \\ \cline{2-3}
          & Quantity of subpopulation & 16 \\ \cline{2-3}
          & Terminal criterion & 2000 migration rounds having been done \\
          \cline{1-3}
  \end{tabular}
\end{table}
This table shows that mutation rate, $p_{mu}$, and mapping rate, $p_{ma}$, change during a run according to Formula~\ref{form:seven} and Formula~ \ref{form:eight}, respectively.
\begin{equation}
 p_{mu} = {p_{mu0}} \times (1-\frac{g_n}{g}\times 0.5)
 \label{form:seven}
\end{equation}
\begin{equation}
p_{ma} = {p_{ma0}} \times (\frac{{{g_n} \times 2}}{g} + 1)
\label{form:eight}
\end{equation}
In Formula~\ref{form:seven}, $p_{mu0}$ is the initial $p_{mu}$.
Similarly, $p_{ma0}$ is the initial $p_{ma}$ in Formula~\ref{form:eight}.
In both formulas, $g_n$ denotes current generations and $g$ represents the maximal ones.

Since the EA used in our experiment can find the optimum solution of many TSP instances smaller than a280, eight much larger instances in \cite{reinelt1991tsplib} from pcb442 to vm1084 are chosen in our experiment.
The results of the traditional algorithm are listed in Table~\ref{table:02}.
Those of the DEA based on our scheme under different value combinations of $\alpha$ and $\beta$ are listed in Table~\ref{table:03}-\ref{table:11}, respectively.
In these tables, each instance corresponds to five migration intervals.
Under each interval, the average of outcomes and the standard deviation of them is given.
Besides, the optimal solution of each instance provided by \cite{reinelt1991tsplib} is listed in each table.
In Table~\ref{table:03}-\ref{table:11}, results having significant difference with those in Table~\ref{table:02} in terms of t-test with 95\% confidence are highlighted by \textbf{\emph{italics and bold}}.

\begin{table}[!htbp]
  \caption{Results of traditional DEA}
  \centering
  \label{table:02}
  \begin{tabular}{|c|c|c|m{1.5cm}<{\centering}|c|}
   \hline
   Instance  & Interval  &\tabincell{c}{Outcomes \\average} &\tabincell{c}{Standard \\deviation} &\tabincell{c}{Optimal \\solution}
     \\ \cline{1-5}
   \multirow{5}*{pcb442}
   &150000	&50939.3	&7.01 & \\ \cline{2-4}
   &200000	&50937.7	&7.84 & \\ \cline{2-4}
   &250000	&50935.0	&24.85 & 50778 \\ \cline{2-4}
   &300000	&50937.5	&7.54 & \\ \cline{2-4}
   &350000	&50934.6	&10.94 & \\ \cline{1-5}
   \multirow{5}*{p654}
   &50000	&34643.4	&1.22 & \\ \cline{2-4}
   &60000	&34643.4	&1.22 & \\ \cline{2-4}
   &70000	&34643.5	&1.57 & 34643 \\ \cline{2-4}
   &80000	&34643.0	&0.00 & \\ \cline{2-4}
   &90000	&34643.1	&0.37 & \\ \cline{1-5}
   \multirow{5}*{d657}
   &200000 	&49092.7	&36.28 & \\ \cline{2-4}
   &250000 	&49084.5	&27.21 & \\ \cline{2-4}
   &300000	&49063.9	&39.07 &48912 \\ \cline{2-4}
   &350000 	&49083.2	&34.98 & \\ \cline{2-4}
   &400000	&49065.3	&35.79 & \\ \cline{1-5}
   \multirow{5}*{u724}
   &150000 	&42143.3	&43.23 & \\ \cline{2-4}
   &200000 	&42105.0	&37.78 & \\ \cline{2-4}
   &250000 	&42086.1	&43.66 &41910 \\ \cline{2-4}
   &300000 	&42085.6	&32.77 & \\ \cline{2-4}
   &350000 	&42068.8	&33.05 & \\ \cline{1-5}
   \multirow{5}*{rat783}
   &150000 	&8829.1	    &11.71 & \\ \cline{2-4}
   &200000 	&8821.8	    &9.11  & \\ \cline{2-4}
   &250000 	&8818.9	    &7.82  &8806 \\ \cline{2-4}
   &300000 	&8815.6	    &5.97  & \\ \cline{2-4}
   &350000 	&8815.6 	&7.50  & \\ \cline{1-5}
   \multirow{5}*{dsj1000}
   &800000 	&18786697.6 &27651.24 & \\ \cline{2-4}	
   &1000000 	&18756550.5 &20746.29 & \\ \cline{2-4}
   &1200000 	&18761931.7 &19044.51 &18659688 \\ \cline{2-4}
   &1400000 	&18762151.6 &24886.79 & \\ \cline{2-4}
   &1600000 	&18757890.0 &21787.65 & \\ \cline{1-5}
   \multirow{5}*{pr1002}
   &500000  &259680.9 	&253.01 & \\ \cline{2-4}
   &600000  &259551.0 	&233.13 & \\ \cline{2-4}
   &700000  &259553.3 	&249.21 & 259045 \\\cline{2-4}
   &800000  &259399.3 	&159.45 & \\ \cline{2-4}
   &900000  &259459.9 	&208.12 & \\ \cline{1-5}
   \multirow{5}*{vm1084}
   &600000  &239785.1 	&204.25 & \\ \cline{2-4}
   &800000  &239773.2 	&177.70 & \\ \cline{2-4}
   &1000000  &239779.5 	&187.50 & 239297 \\\cline{2-4}
   &1200000  &239738.4 	&178.93 & \\ \cline{2-4}
   &1400000  &239739.5 	&188.81 & \\ \cline{1-5}
  \end{tabular}
\end{table}
%
\begin{table}[!htbp]
  \caption{Results of DEA with our scheme when $\alpha=0.5$ and $\beta=0.5$}
  \centering
  \label{table:03}
  \begin{tabular}{|c|c|c|m{1.5cm}<{\centering}|c|}
   \hline
   Instance  & Interval  &\tabincell{c}{Outcomes \\average} &\tabincell{c}{Standard \\deviation} &\tabincell{c}{Optimal \\solution}
   \\ \cline{1-5}
   \multirow{5}*{pcb442}
   &150000	&50936.8	&8.37 & \\ \cline{2-4}
   &200000	&50935.6	&8.11 & \\ \cline{2-4}
   &250000	&50927.8	&24.62 & 50778 \\ \cline{2-4}
   &\textbf{\emph{300000}}	&\textbf{\emph{50924.8}}	&\textbf{\emph{33.10}} & \\ \cline{2-4}
   &350000	&50934.5	&12.86 & \\ \cline{1-5}
   \multirow{5}*{p654}
   &50000	&34643.1	&0.51& \\ \cline{2-4}
   &60000	&34643.1	&0.37 & \\ \cline{2-4}
   &70000	&34643.0	&0.00 & 34643 \\ \cline{2-4}
   &80000	&34643.1	&0.37 & \\ \cline{2-4}
   &90000	&34643.1	&0.37 & \\ \cline{1-5}
   \multirow{5}*{d657}
   &200000 	&49077.9	&32.29 & \\ \cline{2-4}
   &\textbf{\emph{250000}} 	&\textbf{\emph{49057.9}}	&\textbf{\emph{32.71}} & \\ \cline{2-4}
   &300000	&49059.6	&31.23 &48912 \\ \cline{2-4}
   &\textbf{\emph{350000}} 	&\textbf{\emph{49057.0}}	&\textbf{\emph{30.24}} & \\ \cline{2-4}
   &400000	&49059.1	&45.58 & \\ \cline{1-5}
   \multirow{5}*{u724}
   &\textbf{\emph{150000}} 	&\textbf{\emph{42099.2}}	&\textbf{\emph{31.61}} & \\ \cline{2-4}
   &\textbf{\emph{200000}} 	&\textbf{\emph{42074.6}}	&\textbf{\emph{33.18}} & \\ \cline{2-4}
   &250000 	&42073.3	&40.11 &41910 \\ \cline{2-4}
   &\textbf{\emph{300000}} 	&\textbf{\emph{42064.9}}	&\textbf{\emph{23.90}} & \\ \cline{2-4}
   &350000 	&42060.3	&36.93 & \\ \cline{1-5}
   \multirow{5}*{rat783}
   &150000 	&8825.0	    &7.97 & \\ \cline{2-4}
   &\textbf{\emph{200000}} 	&\textbf{\emph{8817.3}}	    &\textbf{\emph{7.62}}  & \\ \cline{2-4}
   &250000 	&8815.9	    &8.28  &8806 \\ \cline{2-4}
   &300000 	&8814.5	    &6.23  & \\ \cline{2-4}
   &350000 	&8814.8 	&5.52  & \\ \cline{1-5}
   \multirow{5}*{dsj1000}
   &\textbf{\emph{800000}}	    &\textbf{\emph{18761927.0}} &\textbf{\emph{25664.25}} & \\ \cline{2-4}	
   &1000000 	&18762398.9 &23733.48 & \\ \cline{2-4}
   &\textbf{\emph{1200000}} 	&\textbf{\emph{18748623.4}} &\textbf{\emph{24506.86}} &18659688 \\ \cline{2-4}
   &\emph{1400000} 	&\emph{18745199.9} &\emph{27584.08} & \\ \cline{2-4}
   &\emph{1600000} 	&\emph{18744510.3} &\emph{23523.72} & \\ \cline{1-5}
   \multirow{5}*{pr1002}
   &500000  &259551.6 	&268.12 & \\ \cline{2-4}
   &600000  &259504.9 	&223.44 & \\ \cline{2-4}
   &\textbf{\emph{700000}}  &\textbf{\emph{259383.7}} 	&\textbf{\emph{172.11}} &259045 \\\cline{2-4}
   &800000  &259375.9 	&233.52 & \\ \cline{2-4}
   &900000  &259387.2 	&239.40 & \\ \cline{1-5}
   \multirow{5}*{vm1084}
   &600000  &239736.0 	&164.15 & \\ \cline{2-4}
   &\textbf{\emph{800000}}  &\textbf{\emph{239676.7}} 	&\textbf{\emph{141.30}} & \\ \cline{2-4}
   &1000000  &239702.1 	&163.49 & 239297 \\\cline{2-4}
   &1200000  &239682.9 	&179.91 & \\ \cline{2-4}
   &\textbf{\emph{1400000}}  &\textbf{\emph{239637.6}} 	&\textbf{\emph{122.91}} & \\ \cline{1-5}
  \end{tabular}
\end{table}
%
\clearpage
\begin{table}[!htbp]
  \centering
  \caption{Results of DEA with our scheme when $\alpha=0.5$ and $\beta=1.0$}
  \label{table:04}
  \begin{tabular}{|c|c|c|m{1.5cm}<{\centering}|c|}
   \hline
   Instance  & Interval  &\tabincell{c}{Outcomes \\average} &\tabincell{c}{Standard \\deviation} &\tabincell{c}{Optimal \\solution}
   \\ \cline{1-5}
   \multirow{5}*{pcb442}
   &\textbf{\emph{150000}}	&\textbf{\emph{50931.7}}	&\textbf{\emph{12.66}} & \\ \cline{2-4}
   &200000	&50934.3	&10.46 & \\ \cline{2-4}
   &250000	&50929.5	&29.45 & 50778 \\ \cline{2-4}
   &\textbf{\emph{300000}}	&\textbf{\emph{50932.7}}	&\textbf{\emph{10.44}} & \\ \cline{2-4}
   &350000	&50927.1	&22.43 & \\ \cline{1-5}
   \multirow{5}*{p654}
   &50000	&34643.0	&0.00 & \\ \cline{2-4}
   &60000	&34643.0	&0.00 & \\ \cline{2-4}
   &70000	&34643.0	&0.00 & 34643 \\ \cline{2-4}
   &80000	&34643.0	&0.00 & \\ \cline{2-4}
   &90000	&34643.0	&0.00 & \\ \cline{1-5}
   \multirow{5}*{d657}
   &\textbf{\emph{200000}} 	&\textbf{\emph{49055.4}}	&\textbf{\emph{35.44}} & \\ \cline{2-4}
   &\textbf{\emph{250000}} 	&\textbf{\emph{49057.3}}	&\textbf{\emph{42.35}} & \\ \cline{2-4}
   &300000	&49048.0	&38.85 &48912 \\ \cline{2-4}
   &\textbf{\emph{350000}} 	&\textbf{\emph{49045.9}}	&\textbf{\emph{37.06}} & \\ \cline{2-4}
   &\textbf{\emph{400000}}	&\textbf{\emph{49036.6}}	&\textbf{\emph{32.54}} & \\ \cline{1-5}
   \multirow{5}*{u724}
   &\textbf{\emph{150000}} 	&\textbf{\emph{42070.7}}	&\textbf{\emph{33.66}} & \\ \cline{2-4}
   &\textbf{\emph{200000}} 	&\textbf{\emph{42067.9}}	&\textbf{\emph{29.89}} & \\ \cline{2-4}
   &\textbf{\emph{250000}} 	&\textbf{\emph{42054.6}}	&\textbf{\emph{31.99}} &41910 \\ \cline{2-4}
   &\textbf{\emph{300000}} 	&\textbf{\emph{42037.9}}	&\textbf{\emph{39.68}} & \\ \cline{2-4}
   &\textbf{\emph{350000}} 	&\textbf{\emph{42036.7}}	&\textbf{\emph{35.86}} & \\ \cline{1-5}
   \multirow{5}*{rat783}
   &\emph{150000} 	&\emph{8822.8}	    &\emph{9.91} & \\ \cline{2-4}
   &\emph{200000} 	&\emph{8815.8}	    &\emph{7.32}  & \\ \cline{2-4}
   &250000 	&8816.2	    &7.01  &8806 \\ \cline{2-4}
   &300000 	&8812.9	    &5.72  & \\ \cline{2-4}
   &350000 	&8812.9 	&8.13  & \\ \cline{1-5}
   \multirow{5}*{dsj1000}
   &\textbf{\emph{800000}} 	    &\textbf{\emph{18756031.3}} &\textbf{\emph{24623.43}} & \\ \cline{2-4}	
   &1000000 	&18748806.2 &26546.89 & \\ \cline{2-4}
   &\textbf{\emph{1200000}} 	&\textbf{\emph{18748989.6}} &\textbf{\emph{22840.60}} &18659688 \\ \cline{2-4}
   &\textbf{\emph{1400000}} 	&\textbf{\emph{18735264.4}} &\textbf{\emph{20742.59}} & \\ \cline{2-4}
   &\textbf{\emph{1600000}} 	&\textbf{\emph{18738518.6}} &\textbf{\emph{26541.26}} & \\ \cline{1-5}
   \multirow{5}*{pr1002}
   &\textbf{\emph{500000}}  &\textbf{\emph{259433.6}} 	&\textbf{\emph{190.88}} & \\ \cline{2-4}
   &\textbf{\emph{600000}}  &\textbf{\emph{259335.0}} 	&\textbf{\emph{165.78}} & \\ \cline{2-4}
   &\textbf{\emph{700000}}  &\textbf{\emph{259303.5}} 	&\textbf{\emph{140.85}} & 259045 \\\cline{2-4}
   &\textbf{\emph{800000}}  &\textbf{\emph{259280.7}} 	&\textbf{\emph{138.09}} & \\ \cline{2-4}
   &\textbf{\emph{900000}}  &\textbf{\emph{259354.0}} 	&\textbf{\emph{199.96}} & \\ \cline{1-5}
   \multirow{5}*{vm1084}
   &\textbf{\emph{600000}}  &\textbf{\emph{239683.8}} 	&\textbf{\emph{162.64}} & \\ \cline{2-4}
   &\textbf{\emph{800000}}  &\textbf{\emph{239665.7}} 	&\textbf{\emph{136.93}} & \\ \cline{2-4}
   &\textbf{\emph{1000000}}  &\textbf{\emph{239671.2}} 	&\textbf{\emph{135.21}} & 239297 \\\cline{2-4}
   &\textbf{\emph{1200000}}  &\textbf{\emph{239638.1}} 	&\textbf{\emph{145.54}} & \\ \cline{2-4}
   &\textbf{\emph{1400000}}  &\textbf{\emph{239623.6}} 	&\textbf{\emph{126.22}} & \\ \cline{1-5}
  \end{tabular}
\end{table}

\begin{table}[!htbp]
  \centering
  \caption{Results of DEA with our scheme when $\alpha=0.5$ and $\beta=2.0$}
  \label{table:05}
  \begin{tabular}{|c|c|c|m{1.5cm}<{\centering}|c|}
   \hline
   Instance  & Interval  &\tabincell{c}{Outcomes \\average} &\tabincell{c}{Standard \\deviation} &\tabincell{c}{Optimal \\solution}
   \\ \cline{1-5}
   \multirow{5}*{pcb442}
   &\textbf{\emph{150000}}	&\textbf{\emph{50932.6}}	&\textbf{\emph{11.09}} & \\ \cline{2-4}
   &\textbf{\emph{200000}}	&\textbf{\emph{50930.4}}	&\textbf{\emph{10.13}} & \\ \cline{2-4}
   &250000	&50926.6	&27.04 & 50778 \\ \cline{2-4}
   &\textbf{\emph{300000}}	&\textbf{\emph{50919.6}}	&\textbf{\emph{32.26}} & \\ \cline{2-4}
   &\textbf{\emph{350000}}	&\textbf{\emph{50922.3}}	&\textbf{\emph{23.90}} & \\ \cline{1-5}
   \multirow{5}*{p654}
   &50000	&34643.0	&0.00 & \\ \cline{2-4}
   &60000	&34643.0	&0.00 & \\ \cline{2-4}
   &70000	&34643.0	&0.00 & 34643 \\ \cline{2-4}
   &80000	&34643.0	&0.00 & \\ \cline{2-4}
   &90000	&34643.0	&0.00 & \\ \cline{1-5}
   \multirow{5}*{d657}
   &\textbf{\emph{200000}} 	&\textbf{\emph{49050.0}} 	&\textbf{\emph{31.05}} & \\ \cline{2-4}
   &\textbf{\emph{250000}} 	&\textbf{\emph{49054.0}} 	&\textbf{\emph{41.66}} & \\ \cline{2-4}
   &300000	&49045.9 	&34.37 &48912 \\ \cline{2-4}
   &\textbf{\emph{350000}} 	&\textbf{\emph{49028.7}} 	&\textbf{\emph{31.89}} & \\ \cline{2-4}
   &\textbf{\emph{400000}}	&\textbf{\emph{49042.3}} 	&\textbf{\emph{36.07}} & \\ \cline{1-5}
   \multirow{5}*{u724}
   &\textbf{\emph{150000}} 	&\textbf{\emph{42072.2}} 	&\textbf{\emph{31.06}} & \\ \cline{2-4}
   &\textbf{\emph{200000}} 	&\textbf{\emph{42045.2}} 	&\textbf{\emph{27.53}} & \\ \cline{2-4}
   &\textbf{\emph{250000}} 	&\textbf{\emph{42045.9}} 	&\textbf{\emph{30.77}} &41910 \\ \cline{2-4}
   &\textbf{\emph{300000}} 	&\textbf{\emph{42032.3}} 	&\textbf{\emph{29.95}} & \\ \cline{2-4}
   &\textbf{\emph{350000}} 	&\textbf{\emph{42026.7}} 	&\textbf{\emph{26.53}} & \\ \cline{1-5}
   \multirow{5}*{rat783}
   &150000 	&8828.1	    &8.83 & \\ \cline{2-4}
   &200000 	&8817.9	    &6.69  & \\ \cline{2-4}
   &\textbf{\emph{250000}} 	&\textbf{\emph{8813.9}}	    &\textbf{\emph{6.77}}  &8806 \\ \cline{2-4}
   &300000 	&8815.0	    &6.24  & \\ \cline{2-4}
   &350000 	&8813.4 	&6.44  & \\ \cline{1-5}
   \multirow{5}*{dsj1000}
   &\textbf{\emph{800000}} 	&\textbf{\emph{18741740.2}} &\textbf{\emph{22220.14}} & \\ \cline{2-4}	
   &\textbf{\emph{1000000}} 	&\textbf{\emph{18739983.0}} &\textbf{\emph{24819.99}} & \\ \cline{2-4}
   &\textbf{\emph{1200000}} 	&\textbf{\emph{18738972.7}} &\textbf{\emph{18942.56}} &18659688 \\ \cline{2-4}
   &\textbf{\emph{1400000}} 	&\textbf{\emph{18724773.5}} &\textbf{\emph{27336.00}} & \\ \cline{2-4}
   &\textbf{\emph{1600000}} 	&\textbf{\emph{18732647.5}} &\textbf{\emph{20957.86}} & \\ \cline{1-5}
   \multirow{5}*{pr1002}
   &\textbf{\emph{500000}}  &\textbf{\emph{259277.2}} 	&\textbf{\emph{136.14}} & \\ \cline{2-4}
   &\textbf{\emph{600000}}  &\textbf{\emph{259402.0}} 	&\textbf{\emph{201.64}} & \\ \cline{2-4}
   &\textbf{\emph{700000}}  &\textbf{\emph{259342.0}} 	&\textbf{\emph{189.56}} & 259045 \\\cline{2-4}
   &\textbf{\emph{800000}}  &\textbf{\emph{259296.1}} 	&\textbf{\emph{180.98}} & \\ \cline{2-4}
   &\textbf{\emph{900000}}  &\textbf{\emph{259288.9}} 	&\textbf{\emph{132.95}} & \\ \cline{1-5}
   \multirow{5}*{vm1084}
   &600000  &239691.4 	&169.62 & \\ \cline{2-4}
   &800000  &239690.1 	&167.67 & \\ \cline{2-4}
   &\textbf{\emph{1000000}}  &\textbf{\emph{239637.8}} 	&\textbf{\emph{128.26}} & 239297 \\\cline{2-4}
   &\textbf{\emph{1200000}}  &\textbf{\emph{239638.7}} 	&\textbf{\emph{171.76}} & \\ \cline{2-4}
   &\textbf{\emph{1400000}}  &\textbf{\emph{239585.6}} 	&\textbf{\emph{133.70}} & \\ \cline{1-5}
  \end{tabular}
\end{table}
%
\begin{table}[!htbp]
  \centering
  \caption{Results of DEA with our scheme when $\alpha=1.0$ and $\beta=0.5$}
  \label{table:06}
  \begin{tabular}{|c|c|c|m{1.5cm}<{\centering}|c|}
   \hline
   Instance  & Interval  &\tabincell{c}{Outcomes \\average} &\tabincell{c}{Standard \\deviation} &\tabincell{c}{Optimal \\solution}
   \\ \cline{1-5}
   \multirow{5}*{pcb442}
   &150000	&50938.3 	&6.86 & \\ \cline{2-4}
   &200000	&50937.4 	&8.53 & \\ \cline{2-4}
   &250000	&50934.7 	&10.22 & 50778 \\ \cline{2-4}
   &300000	&50933.5 	&27.43 & \\ \cline{2-4}
   &350000	&50921.9 	&35.34 & \\ \cline{1-5}
   \multirow{5}*{p654}
   &50000	&34643.3	&1.18 & \\ \cline{2-4}
   &60000	&34643.6	&2.30 & \\ \cline{2-4}
   &70000	&34643.1	&0.37 & 34643 \\ \cline{2-4}
   &80000	&34643.0	&0.00 & \\ \cline{2-4}
   &90000	&34643.0	&0.00 & \\ \cline{1-5}
   \multirow{5}*{d657}
   &200000 	&49076.7 	&42.28 & \\ \cline{2-4}
   &250000 	&49083.5 	&33.96 & \\ \cline{2-4}
   &300000	&49070.3 	&32.63 &48912 \\ \cline{2-4}
   &\textbf{\emph{350000}} 	&\textbf{\emph{49060.2}} 	&\textbf{\emph{29.56}} & \\ \cline{2-4}
   &400000	&49058.7 	&42.51 & \\ \cline{1-5}
   \multirow{5}*{u724}
   &\textbf{\emph{150000}} 	&\textbf{\emph{42093.8}} 	&\textbf{\emph{39.47}} & \\ \cline{2-4}
   &\textbf{\emph{200000}} 	&\textbf{\emph{42081.5}} 	&\textbf{\emph{31.40}} & \\ \cline{2-4}
   &250000 	&42081.8 	&30.52 &41910 \\ \cline{2-4}
   &\textbf{\emph{300000}} 	&\textbf{\emph{42064.9}} 	&\textbf{\emph{28.86}} & \\ \cline{2-4}
   &350000 	&42071.3 	&30.47 & \\ \cline{1-5}
   \multirow{5}*{rat783}
   &150000 	&8825.9 	&10.50 & \\ \cline{2-4}
   &200000 	&8818.5 	&8.52  & \\ \cline{2-4}
   &250000 	&8816.4 	&6.64  &8806 \\ \cline{2-4}
   &300000 	&8814.4 	&6.72  & \\ \cline{2-4}
   &350000 	&8814.0 	&7.31  & \\ \cline{1-5}
   \multirow{5}*{dsj1000}
   &\textbf{\emph{800000}} 	&\textbf{\emph{18773122.2}} 	&\textbf{\emph{23162.37}} & \\ \cline{2-4}	
   &1000000 	&18761402.1 	&26171.98 & \\ \cline{2-4}
   &1200000 	&18755819.7 	&15957.70 &18659688 \\ \cline{2-4}
   &\textbf{\emph{1400000}} 	&\textbf{\emph{18746441.0}} 	&\textbf{\emph{23247.20}} & \\ \cline{2-4}
   &1600000 	&18752288.1 	&19447.99 & \\ \cline{1-5}
   \multirow{5}*{pr1002}
   &\textbf{\emph{500000}}  &\textbf{\emph{259517.0}}	&\textbf{\emph{273.21}} & \\ \cline{2-4}
   &600000  &259468.7 	&236.99 & \\ \cline{2-4}
   &700000  &259449.1 	&199.82 & 259045 \\\cline{2-4}
   &\textbf{\emph{800000}}  &\textbf{\emph{259515.5}} 	&\textbf{\emph{272.15}} & \\ \cline{2-4}
   &900000  &259499.3 	&226.40 & \\ \cline{1-5}
   \multirow{5}*{vm1084}
   &600000  &239770.3 	&162.83 & \\ \cline{2-4}
   &800000  &239730.6 	&137.43 & \\ \cline{2-4}
   &1000000  &239693.5 	&151.11& 239297 \\\cline{2-4}
   &1200000  &239694.1 	&158.66 & \\ \cline{2-4}
   &1400000  &239670.1 	&132.08 & \\ \cline{1-5}
  \end{tabular}
\end{table}

\begin{table}[!htbp]
  \centering
  \caption{Results of DEA with our scheme when $\alpha=1.0$ and $\beta=1.0$}
  \label{table:07}
  \begin{tabular}{|c|c|c|m{1.5cm}<{\centering}|c|}
   \hline
   Instance  & Interval  &\tabincell{c}{Outcomes \\average} &\tabincell{c}{Standard \\deviation} &\tabincell{c}{Optimal \\solution}
   \\ \cline{1-5}
   \multirow{5}*{pcb442}
   &150000	&50931.5	&20.76 & \\ \cline{2-4}
   &200000	&50929.5	&26.70 & \\ \cline{2-4}
   &250000	&50933.9	&12.78 & 50778 \\ \cline{2-4}
   &\textbf{\emph{300000}}	&\textbf{\emph{50931.7}}	&\textbf{\emph{10.29}} & \\ \cline{2-4}
   &350000	&50931.2	&22.20 & \\ \cline{1-5}
   \multirow{5}*{p654}
   &50000	&34643.1	&0.37 & \\ \cline{2-4}
   &60000	&34643.1	&0.37 & \\ \cline{2-4}
   &70000	&34643.4	&1.52 & 34643 \\ \cline{2-4}
   &80000	&34643.0	&0.00 & \\ \cline{2-4}
   &90000	&34643.0	&0.00 & \\ \cline{1-5}
   \multirow{5}*{d657}
   &\textbf{\emph{200000}} 	&\textbf{\emph{49067.4}}	&\textbf{\emph{43.30}} & \\ \cline{2-4}
   &\textbf{\emph{250000}} 	&\textbf{\emph{49063.7}}	&\textbf{\emph{37.62}} & \\ \cline{2-4}
   &300000	&49055.5	&36.67 &48912 \\ \cline{2-4}
   &\textbf{\emph{350000}} 	&\textbf{\emph{49050.3}}	&\textbf{\emph{32.32}} & \\ \cline{2-4}
   &400000	&49046.9	&43.36 & \\ \cline{1-5}
   \multirow{5}*{u724}
   &\textbf{\emph{150000}} 	&\textbf{\emph{42094.9}} 	&\textbf{\emph{29.82}} & \\ \cline{2-4}
   &\textbf{\emph{200000}} 	&\textbf{\emph{42064.9}} 	&\textbf{\emph{42.94}} & \\ \cline{2-4}
   &250000 	&42086.1	&42067.8 	&34.59 \\ \cline{2-4}
   &\textbf{\emph{300000}} 	&\textbf{\emph{42066.3}} 	&\textbf{\emph{25.19}} & \\ \cline{2-4}
   &350000 	&42054.6 	&30.69 & \\ \cline{1-5}
   \multirow{5}*{rat783}
   &150000 	&8824.0 	&8.83 & \\ \cline{2-4}
   &200000 	&8819.7 	&7.47 & \\ \cline{2-4}
   &\textbf{\emph{250000}} 	&\textbf{\emph{8815.1}} 	&\textbf{\emph{6.77}} &8806 \\ \cline{2-4}
   &300000 	&8814.0 	&7.05 & \\ \cline{2-4}
   &350000 	&8812.5 	&6.69 & \\ \cline{1-5}
   \multirow{5}*{dsj1000}
   &\textbf{\emph{800000}} 	&\textbf{\emph{18763625.5}} &\textbf{\emph{21831.34}} & \\ \cline{2-4}	
   &1000000 	&18762785.8 &24970.92 & \\ \cline{2-4}
   &\textbf{\emph{1200000}} 	&\textbf{\emph{18746381.2}} &\textbf{\emph{20736.39}} &18659688 \\ \cline{2-4}
   &\textbf{\emph{1400000}}	&\textbf{\emph{18745791.7}} &\textbf{\emph{17386.04}} & \\ \cline{2-4}
   &\textbf{\emph{1600000}} 	&\textbf{\emph{18743302.0}} &\textbf{\emph{18939.43}} & \\ \cline{1-5}
   \multirow{5}*{pr1002}
   &\textbf{\emph{500000}}  &\textbf{\emph{259435.1}} 	&\textbf{\emph{224.69}} & \\ \cline{2-4}
   &\textbf{\emph{600000}}  &\textbf{\emph{259417.3}} 	&\textbf{\emph{222.38}} & \\ \cline{2-4}
   &\textbf{\emph{700000}}  &\textbf{\emph{259359.6}} 	&\textbf{\emph{169.53}} & 259045 \\\cline{2-4}
   &800000  &259341.7 	&198.52 & \\ \cline{2-4}
   &\textbf{\emph{900000}}  &\textbf{\emph{259317.7}} 	&\textbf{\emph{187.90}} & \\ \cline{1-5}
   \multirow{5}*{vm1084}
   &600000  &239739.4 	&183.33 & \\ \cline{2-4}
   &800000  &239694.4 	&158.19 & \\ \cline{2-4}
   &1000000  &239691.5 	&168.34 & 239297 \\\cline{2-4}
   &1200000  &239699.0 	&143.17 & \\ \cline{2-4}
   &1400000  &239668.0 	&129.28 & \\ \cline{1-5}
  \end{tabular}
\end{table}
\begin{table}[!htbp]
  \centering
  \caption{Results of DEA with our scheme when $\alpha=1.0$ and $\beta=2.0$}
  \label{table:08}
  \begin{tabular}{|c|c|c|m{1.5cm}<{\centering}|c|}
   \hline
   Instance  & Interval  &\tabincell{c}{Outcomes \\average} &\tabincell{c}{Standard \\deviation} &\tabincell{c}{Optimal \\solution}
   \\ \cline{1-5}
   \multirow{5}*{pcb442}
   &\textbf{\emph{150000}}	&\textbf{\emph{50934.0}} 	&\textbf{\emph{10.64}} & \\ \cline{2-4}
   &\textbf{\emph{200000}}	&\textbf{\emph{50931.3}} 	&\textbf{\emph{11.15}} & \\ \cline{2-4}
   &250000	&50932.7 	&12.33 & 50778 \\ \cline{2-4}
   &\textbf{\emph{300000}}	&\textbf{\emph{50931.5}} 	&\textbf{\emph{10.96}} & \\ \cline{2-4}
   &\textbf{\emph{350000}}	&\textbf{\emph{50927.6}} 	&\textbf{\emph{14.41}} & \\ \cline{1-5}
   \multirow{5}*{p654}
   &50000	&34643.0	&0.00 & \\ \cline{2-4}
   &60000	&34643.0	&0.00 & \\ \cline{2-4}
   &70000	&34643.0	&0.00 & 34643 \\ \cline{2-4}
   &80000	&34643.1	&0.37 & \\ \cline{2-4}
   &90000	&34643.0	&0.00 & \\ \cline{1-5}
   \multirow{5}*{d657}
   &\textbf{\emph{200000}} 	&\textbf{\emph{49052.7}} 	&\textbf{\emph{35.84}} & \\ \cline{2-4}
   &\textbf{\emph{250000}}	&\textbf{\emph{49053.1}} 	&\textbf{\emph{36.05}} & \\ \cline{2-4}
   &300000	&49050.2 	&34.89 &48912 \\ \cline{2-4}
   &\textbf{\emph{350000}} 	&\textbf{\emph{49031.9}} 	&\textbf{\emph{34.44}} & \\ \cline{2-4}
   &\textbf{\emph{400000}}	&\textbf{\emph{49041.6}} 	&\textbf{\emph{47.01}} & \\ \cline{1-5}
   \multirow{5}*{u724}
   &\textbf{\emph{150000}} 	&\textbf{\emph{42070.6}} 	&\textbf{\emph{37.78}} & \\ \cline{2-4}
   &\textbf{\emph{200000}} 	&\textbf{\emph{42056.0}} 	&\textbf{\emph{37.16}} & \\ \cline{2-4}
   &\textbf{\emph{250000}} 	&\textbf{\emph{42035.9}} 	&\textbf{\emph{27.50}} &41910 \\ \cline{2-4}
   &\textbf{\emph{300000}} 	&\textbf{\emph{42031.8}} 	&\textbf{\emph{30.09}} & \\ \cline{2-4}
   &\textbf{\emph{350000}} 	&\textbf{\emph{42033.0}} 	&\textbf{\emph{30.42}} & \\ \cline{1-5}
   \multirow{5}*{rat783}
   &150000 	&8824.6 	&9.78 & \\ \cline{2-4}
   &200000 	&8817.7 	&8.45 & \\ \cline{2-4}
   &\textbf{\emph{250000}} 	&\textbf{\emph{8814.6}} 	&\textbf{\emph{6.62}}  &8806 \\ \cline{2-4}
   &300000 	&8812.7 	&6.50  & \\ \cline{2-4}
   &350000 	&8814.3 	&6.14  & \\ \cline{1-5}
   \multirow{5}*{dsj1000}
   &\textbf{\emph{800000}} 	&\textbf{\emph{18740584.7}} 	&\textbf{\emph{20775.90}} & \\ \cline{2-4}	
   &\textbf{\emph{1000000}} 	&\textbf{\emph{18743355.3}} 	&\textbf{\emph{20831.26}} & \\ \cline{2-4}
   &\textbf{\emph{1200000}} 	&\textbf{\emph{18741427.6}} 	&\textbf{\emph{24245.91}} &18659688 \\ \cline{2-4}
   &\textbf{\emph{1400000}} 	&\textbf{\emph{18729556.2}} 	&\textbf{\emph{19250.64}} & \\ \cline{2-4}
   &\textbf{\emph{1600000}} 	&\textbf{\emph{18733965.3}} 	&\textbf{\emph{25656.26}}
 & \\ \cline{1-5}
   \multirow{5}*{pr1002}
   &\textbf{\emph{500000}}  &\textbf{\emph{259394.7}} 	&\textbf{\emph{204.43}} & \\ \cline{2-4}
   &\textbf{\emph{600000}}  &\textbf{\emph{259367.8}} 	&\textbf{\emph{198.61}} & \\ \cline{2-4}
   &\textbf{\emph{700000}}  &\textbf{\emph{259331.4}} 	&\textbf{\emph{180.84}} & 259045 \\\cline{2-4}
   &800000  &259410.4 	&179.61 & \\ \cline{2-4}
   &\textbf{\emph{900000}}  &\textbf{\emph{259307.4}} 	&\textbf{\emph{181.98}} & \\ \cline{1-5}
   \multirow{5}*{vm1084}
   &600000  &239724.6 	&145.35 & \\ \cline{2-4}
   &800000  &239717.7 	&149.77 & \\ \cline{2-4}
   &\textbf{\emph{1000000}}  &\textbf{\emph{239638.4}} 	&\textbf{\emph{148.45}} & 239297 \\\cline{2-4}
   &\textbf{\emph{1200000}}  &\textbf{\emph{239642.8}} 	&\textbf{\emph{116.59}} & \\ \cline{2-4}
   &\textbf{\emph{1400000}}  &\textbf{\emph{239644.8}} 	&\textbf{\emph{142.91}} & \\ \cline{1-5}
  \end{tabular}
\end{table}
\begin{table}[!htbp]
  \centering
  \caption{Results of DEA with our scheme when $\alpha=2.0$ and $\beta=0.5$}
  \label{table:09}
  \begin{tabular}{|c|c|c|m{1.5cm}<{\centering}|c|}
   \hline
   Instance  & Interval  &\tabincell{c}{Outcomes \\average} &\tabincell{c}{Standard \\deviation} &\tabincell{c}{Optimal \\solution}
   \\ \cline{1-5}
   \multirow{5}*{pcb442}
   &\textbf{\emph{150000}}	&\textbf{\emph{50933.8}} 	&\textbf{\emph{9.27}} & \\ \cline{2-4}
   &200000	&50936.4 	&8.19 & \\ \cline{2-4}
   &250000	&50934.8 	&10.53 & 50778 \\ \cline{2-4}
   &300000	&50936.1 	&9.49 & \\ \cline{2-4}
   &350000	&50937.4 	&8.42 & \\ \cline{1-5}
   \multirow{5}*{p654}
   &50000	&34643.4	&1.52 & \\ \cline{2-4}
   &60000	&34643.1	&0.37 & \\ \cline{2-4}
   &70000	&34643.1	&0.51 & 34643 \\ \cline{2-4}
   &80000	&34643.3	&1.14 & \\ \cline{2-4}
   &90000	&34643.0	&0.00 & \\ \cline{1-5}
   \multirow{5}*{d657}
   &200000 	&49085.9 	&38.80 & \\ \cline{2-4}
   &250000 	&49095.7 	&30.77 & \\ \cline{2-4}
   &300000	&49075.7 	&29.14 &48912 \\ \cline{2-4}
   &350000 	&49065.2 	&40.71 & \\ \cline{2-4}
   &400000	&49066.2 	&32.46 & \\ \cline{1-5}
   \multirow{5}*{u724}
   &\textbf{\emph{150000}} 	&\textbf{\emph{42113.1}} 	&\textbf{\emph{41.00}} & \\ \cline{2-4}
   &200000 	&42097.8 	&41.37 & \\ \cline{2-4}
   &250000 	&42084.7 	&43.12 &41910 \\ \cline{2-4}
   &300000 	&42076.8 	&30.95 & \\ \cline{2-4}
   &350000 	&42077.5 	&37.93 & \\ \cline{1-5}
   \multirow{5}*{rat783}
   &150000 	&8828.0 	&8.28 & \\ \cline{2-4}
   &200000 	&8818.5 	&6.06  & \\ \cline{2-4}
   &\textbf{\emph{250000}} 	&\textbf{\emph{8814.8}} 	&\textbf{\emph{6.23}}  &8806 \\ \cline{2-4}
   &300000 	&8816.0 	&5.43  & \\ \cline{2-4}
   &350000 	&8814.6 	&5.70  & \\ \cline{1-5}
   \multirow{5}*{dsj1000}
   &800000 	    &18776618.6 	&27998.28 & \\ \cline{2-4}	
   &1000000 	&18761359.1 	&26059.44 & \\ \cline{2-4}
   &1200000 	&18753840.4 	&22750.80 &18659688 \\ \cline{2-4}
   &1400000 	&18753787.1 	&26156.22 & \\ \cline{2-4}
   &1600000 	&18752323.7 	&21359.45 & \\ \cline{1-5}
   \multirow{5}*{pr1002}
   &500000  &259622.7 	&228.65 & \\ \cline{2-4}
   &600000  &259523.0 	&314.84 & \\ \cline{2-4}
   &700000  &259479.1 	&211.93 & 259045 \\\cline{2-4}
   &\textbf{\emph{800000}}  &\textbf{\emph{259500.5}} 	&\textbf{\emph{208.99}} & \\ \cline{2-4}
   &900000  &259519.0 	&196.58 & \\ \cline{1-5}
   \multirow{5}*{vm1084}
   &600000  &239765.1 	&122.05 & \\ \cline{2-4}
   &800000  &239744.2 	&134.48 & \\ \cline{2-4}
   &\textbf{\emph{1000000}}  &\textbf{\emph{239774.2}} 	&\textbf{\emph{153.76}} & 239297 \\\cline{2-4}
   &1200000  &239675.7 	&164.24 & \\ \cline{2-4}
   &1400000  &239661.6 	&125.02 & \\ \cline{1-5}
  \end{tabular}
\end{table}
\begin{table}[!htbp]
  \centering
  \caption{Results of DEA with our scheme when $\alpha=2.0$ and $\beta=1.0$}
  \label{table:10}
  \begin{tabular}{|c|c|c|m{1.5cm}<{\centering}|c|}
   \hline
   Instance  & Interval  &\tabincell{c}{Outcomes \\average} &\tabincell{c}{Standard \\deviation} &\tabincell{c}{Optimal \\solution}
   \\ \cline{1-5}
   \multirow{5}*{pcb442}
   &\textbf{\emph{150000}}	&\textbf{\emph{50934.5}} 	&\textbf{\emph{9.73}} & \\ \cline{2-4}
   &200000	&50935.6 	&25.76 & \\ \cline{2-4}
   &250000	&50937.1 	&8.39 & 50778 \\ \cline{2-4}
   &300000	&50935.8 	&8.25 & \\ \cline{2-4}
   &350000	&50935.2 	&11.55 & \\ \cline{1-5}
   \multirow{5}*{p654}
   &50000	&34643.3	&0.69 & \\ \cline{2-4}
   &60000	&34643.1	&0.37 & \\ \cline{2-4}
   &70000	&34643.0	&0.00 & 34643 \\ \cline{2-4}
   &80000	&34643.0	&0.00 & \\ \cline{2-4}
   &90000	&34643.1	&0.37 & \\ \cline{1-5}
   \multirow{5}*{d657}
   &200000 	&49078.2 	&42.55 & \\ \cline{2-4}
   &250000 	&49078.7 	&47.63 & \\ \cline{2-4}
   &300000	&49063.7 	&42.80 &48912 \\ \cline{2-4}
   &350000 	&49068.5 	&41.41 & \\ \cline{2-4}
   &400000	&49065.7 	&40.75 & \\ \cline{1-5}
   \multirow{5}*{u724}
   &\textbf{\emph{150000}} 	&\textbf{\emph{42112.0}} 	&\textbf{\emph{30.64}} & \\ \cline{2-4}
   &\textbf{\emph{200000}} 	&\textbf{\emph{42081.4}} 	&\textbf{\emph{26.93}} & \\ \cline{2-4}
   &250000 	&42080.1 	&38.89 &41910 \\ \cline{2-4}
   &300000 	&42069.0 	&38.67 & \\ \cline{2-4}
   &350000 	&42064.8 	&27.71 & \\ \cline{1-5}
   \multirow{5}*{rat783}
   &150000 	&8827.5 	&8.08 & \\ \cline{2-4}
   &\textbf{\emph{200000}} 	&\textbf{\emph{8817.2}} 	&\textbf{\emph{4.93}}  & \\ \cline{2-4}
   &250000 	&8816.7 	&8.42  &8806 \\ \cline{2-4}
   &300000 	&8816.5 	&6.18  & \\ \cline{2-4}
   &350000 	&8815.4 	&5.77  & \\ \cline{1-5}
   \multirow{5}*{dsj1000}
   &\textbf{\emph{800000}} 	&\textbf{\emph{18762595.7}} 	&\textbf{\emph{30773.90}} & \\ \cline{2-4}	
   &1000000 	&18765975.2 	&33663.58 & \\ \cline{2-4}
   &1200000 	&18759551.7 	&22376.08 &18659688 \\ \cline{2-4}
   &1400000 	&18755129.5 	&24348.98 & \\ \cline{2-4}
   &1600000 	&18753479.1 	&21177.11 & \\ \cline{1-5}
   \multirow{5}*{pr1002}
   &500000  &259586.6 	&235.77 & \\ \cline{2-4}
   &600000  &259520.7 	&222.06 & \\ \cline{2-4}
   &700000  &259454.7 	&244.08 & 259045 \\\cline{2-4}
   &800000  &259351.2 	&196.07 & \\ \cline{2-4}
   &900000  &259401.3 	&166.26 & \\ \cline{1-5}
   \multirow{5}*{vm1084}
   &600000  &239746.5 	&145.48 & \\ \cline{2-4}
   &800000  &239690.4 	&160.28 & \\ \cline{2-4}
   &1000000  &239742.5 	&204.31 & 239297 \\\cline{2-4}
   &1200000  &239709.6 	&160.58 & \\ \cline{2-4}
   &1400000  &239715.3 	&155.95 & \\ \cline{1-5}
  \end{tabular}
\end{table}
\begin{table}[!htbp]
  \centering
  \caption{Results of DEA with our scheme when $\alpha=2.0$ and $\beta=2.0$}
  \label{table:11}
  \begin{tabular}{|c|c|c|m{1.5cm}<{\centering}|c|}
   \hline
   Instance  & Interval  &\tabincell{c}{Outcomes \\average} &\tabincell{c}{Standard \\deviation} &\tabincell{c}{Optimal \\solution}
   \\ \cline{1-5}
   \multirow{5}*{pcb442}
   &150000	&50937.6 	&8.36 & \\ \cline{2-4}
   &200000	&50938.5 	&8.21 & \\ \cline{2-4}
   &250000	&50935.4 	&10.23 & 50778 \\ \cline{2-4}
   &300000	&50934.7 	&10.39 & \\ \cline{2-4}
   &350000	&50924.3 	&31.84 & \\ \cline{1-5}
   \multirow{5}*{p654}
   &50000	&34643.1	&0.51 & \\ \cline{2-4}
   &60000	&34643.0	&0.00 & \\ \cline{2-4}
   &70000	&34643.0	&0.00 & 34643 \\ \cline{2-4}
   &80000	&34643.0	&0.00 & \\ \cline{2-4}
   &90000	&34643.0	&0.00 & \\ \cline{1-5}
   \multirow{5}*{d657}
   &\textbf{\emph{200000}} 	&\textbf{\emph{49060.4}} 	&\textbf{\emph{42.18}} & \\ \cline{2-4}
   &\textbf{\emph{250000}} 	&\textbf{\emph{49054.0}} 	&\textbf{\emph{35.61}} & \\ \cline{2-4}
   &300000	&49046.5 	&39.59 &48912 \\ \cline{2-4}
   &\textbf{\emph{350000}} 	&\textbf{\emph{49050.9}} 	&\textbf{\emph{30.57}} & \\ \cline{2-4}
   &400000	&49052.1 	&35.16 & \\ \cline{1-5}
   \multirow{5}*{u724}
   &\textbf{\emph{150000}} 	&\textbf{\emph{42080.9}} 	&\textbf{\emph{27.08}} & \\ \cline{2-4}
   &200000 	&42089.8 	&35.98 & \\ \cline{2-4}
   &250000 	&42078.1 	&35.95 &41910 \\ \cline{2-4}
   &300000 	&42071.6 	&42.63 & \\ \cline{2-4}
   &350000 	&42052.2 	&39.33 & \\ \cline{1-5}
   \multirow{5}*{rat783}
   &\textbf{\emph{150000}} 	&\textbf{\emph{8821.6}} 	&\textbf{\emph{9.10}} & \\ \cline{2-4}
   &200000 	&8819.5 	&8.56  & \\ \cline{2-4}
   &250000 	&8816.2 	&6.56  &8806 \\ \cline{2-4}
   &300000 	&8815.4 	&6.30  & \\ \cline{2-4}
   &350000 	&8814.8 	&5.12  & \\ \cline{1-5}
   \multirow{5}*{dsj1000}
   &\textbf{\emph{800000}} 	&\textbf{\emph{18761976.8}} 	&\textbf{\emph{25253.82}} & \\ \cline{2-4}	
   &1000000 	&18760042.3 	&14768.00 & \\ \cline{2-4}
   &1200000 	&18755479.9 	&22505.88 &18659688 \\ \cline{2-4}
   &\textbf{\emph{1400000}} 	&\textbf{\emph{18748354.6}} 	&\textbf{\emph{22603.43}} & \\ \cline{2-4}
   &1600000 	&18746204.2 	&23893.81 & \\ \cline{1-5}
   \multirow{5}*{pr1002}
   &500000  &259537.5 	&306.89 & \\ \cline{2-4}
   &\textbf{\emph{600000}}  &\textbf{\emph{259412.2}} 	&\textbf{\emph{244.67}} & \\ \cline{2-4}
   &700000  &259464.5 	&237.79 & 259045 \\\cline{2-4}
   &800000  &259410.7 	&225.29 & \\ \cline{2-4}
   &900000  &259372.2 	&182.52 & \\ \cline{1-5}
   \multirow{5}*{vm1084}
   &600000  &239789.1 	&252.07 & \\ \cline{2-4}
   &800000  &239706.1 	&172.80 & \\ \cline{2-4}
   &1000000  &239746.1 	&168.84 & 239297 \\\cline{2-4}
   &\textbf{\emph{1200000}}  &\textbf{\emph{239653.7}} 	&\textbf{\emph{146.65}} & \\ \cline{2-4}
   &1400000  &239660.9 	&140.98 & \\ \cline{1-5}
  \end{tabular}
\end{table}
\clearpage

Based on the results in Table~\ref{table:02}-\ref{table:11}, the difficulty of a instance to the two algorithms, DF, can be expressed according to Formula~\ref{form:nine}, where $f_{ave}$ denotes the average of all solutions for one instance obtained in our experiment and $f_o$ represents the optimal solution provided by~\cite{reinelt1991tsplib}.
\begin{equation}
 DF = \frac{f_{ave}-f_o}{f_o}
\label{form:nine}
\end{equation}

The value of difficulty is listed in Table~\ref{table:12}.
\begin{table}[!htbp]
  \centering
  \caption{Difficulty of each instance for DEAs}
  \label{table:12}
  \begin{tabular}{|c|c|c|c|c|}
     \hline
   Instance &pcb442	&p654	&d657	&u724	\\
    \hline
    Difficulty &0.003048 	&0.000003 	&0.003040 	&0.003842  \\
    \hline
   Instance &rat783	&dsj1000	&pr1002	&vm1084 \\
    \hline
    Difficulty &0.001337 	&0.004947 	&0.001482 	&0.001682 \\
     \hline
  \end{tabular}
\end{table}
According to Table~\ref{table:12}, the difficulty of each instance can be classified into three levels.
That of p654 belongs to the lowest level.
Rat783, pr1002 and vm1084 have an intermediate level difficulty.
The rest four instances, pcb442, d657, u724, dsj1000, are ones with high difficulty.

It can be seen in Table~\ref{table:02}-\ref{table:11} that, for the lowest difficulty instance, p654, significant difference between solutions of the DEA based on our scheme and those of the traditional one can be found in no case.
For the intermediate difficulty instances, the DEA with the proposed scheme significantly wins in forty-five cases out of one hundred and thirty-five ones (45/135) and statistically loses in two cases (2/135).
It should be noted that, for high difficulty instances, the DEA with our scheme yields significantly better outcomes than its peer in ninety cases out of one hundred and eighty ones (90/180).
Meanwhile, there are no significant differences in all the rest cases.
Also, the tables show that the solutions' standard deviation of the DEA with our scheme is less than that of the traditional one in two hundred and twenty-six cases out of the all ones (226/360).

Moreover, it can been seen that, under different value combinations of $\alpha$ and $\beta$, the performance of the algorithm with our scheme is significant different.
When $\alpha=0.5$, $\beta=1.0$ or $\alpha=0.5$, $\beta=2.0$, for the all eight instances, the algorithm significantly wins in the most cases (27/40) and never statistically loses.
When $\alpha=1.0$, $\beta=2.0$, the DEA with our scheme also has a good behavior.
In detail, it significantly wins in twenty-six cases (26/40) and never statistically loses.
On the whole, the winning rate is 135/360, while the losing rate is 2/360.
Besides, under the three outstanding value combinations, solutions of the DEA with our scheme have a better standard deviation in eighty-two cases out of one hundred and twenty ones (82/120).
This rate is better than that on the whole (226/360).
Fig.~\ref{figure:01}-\ref{figure:09} show that, under the three outstanding value combinations, the graph of the function in Formula~\ref{eq:04} have the same characteristic which is distinguished from that under the other six ones.

In conclusion, results shows that the DEA based on our scheme has an advantage on solutions.
Moreover, a value combination of $\alpha$ and $\beta$ which makes the graphing slope of the function in Formula~\ref{eq:04} increase monotonically from $-\infty$ to zero when $d\in [0,1]$ is fit for our scheme.

\section{Conclusion}
In this paper, we have presented the scheme of setting the success rate of migration based on subpopulation diversity at each interval for DEAs.
Under the control of the scheme, immigrants enter the target subpopulation at a probability, which is the function of subpopulation diversity according to Formula~\ref{eq:04}, at intervals.
In our experiment, eight instances of the TSP are used to test algorithms.
In detail, under nine different value combinations of parameters required for Formula~\ref{eq:04}, outcomes of the algorithm with the scheme are compared with those of the traditional DEA, respectively.
The experimental results show that, especially for high difficulty instances, the DEA based on our scheme has a significant advantage on solutions.
Moreover, under three value combinations of the parameters making graphing slope of the function in Formula~\ref{eq:04} increase monotonically from $-\infty$ to zero when $d\in [0,1]$, the algorithm with the scheme has most outstanding performance.

Based on the key factor in the behavior of evolutionary computation, diversity, we propose this scheme and discuss its parameters setting.
This scheme can be used to improve solutions of DEAs for diversified problems.
To apply the scheme, first of all, finding the diversity measure of a certain chromosome coding is necessary.
In future, how to use this scheme together with some other diversity based methods should be studied to further improve DEAs.
Also, the simplified calculation method for subpopulation diversity to reduce the time complexity used in our scheme should be discussed in theory.

\section*{Acknowledgment}
The authors would like to thank Dr. Dunhui Xiao and assistant researcher Jian Wang for their valuable suggestions.

\ifCLASSOPTIONcaptionsoff
  \newpage
\fi

\bibliographystyle{IEEEtran}
\bibliography{IEEEabrv,calling}





%
%
%
\end{document}